\documentclass[11pt,bezier,amstex]{article}

\usepackage[in]{fullpage}
\usepackage{authblk}
\usepackage[utf8]{inputenc} 
\usepackage[T1]{fontenc}    
\usepackage{hyperref}       
\usepackage{url}            
\usepackage{booktabs}       
\usepackage{amsfonts}       
\usepackage{nicefrac} 
\usepackage{graphicx}
\usepackage{subfigure}
\usepackage{amsmath,amssymb}
\usepackage{tabularx}
\usepackage{enumitem}
\usepackage{hhline}
\usepackage{makecell}
\usepackage{multirow}
\usepackage{multicol}

\newcommand{\beq}{\begin{equation}}
\newcommand{\eeq}{\end{equation}}

\newcommand\R{\mathbb{R}}

\newcommand{\y}{\mathbf{y}}

\newcommand{\vertiii}[1]{{\left\vert\kern-0.25ex\left\vert\kern-0.25ex\left\vert #1
    \right\vert\kern-0.25ex\right\vert\kern-0.25ex\right\vert}}

\newcommand{\norm}[2]{\ensuremath \|#1\|_{#2}}

\DeclareMathOperator{\argmin}{argmin}

\newcommand {\commentout}[1] {}

\def\ints{{{\rm Z} \kern -.35em {\rm Z} }}  
\def\smallints{{{\rm Z} \kern -.3em {\rm Z} }}  
\def\pints{{{\rm I} \kern -.15em {\rm N} }}      
\newcommand{\reals}{\mathbb R}

\def\cplx{{{\rm I} \kern -.45em {\rm C} }}       
\def\l2{\rm {\mathcal L}^{2}(\reals)}            

\renewcommand{\norm}[1]{\lVert#1\rVert}

\newcommand{\be}{\begin{eqnarray}}
\newcommand{\ee}{\end{eqnarray}}
\newcommand{\bea}{\begin{eqnarray}}
\newcommand{\eea}{\end{eqnarray}}
\newcommand{\beaa}{\begin{eqnarray*}}
\newcommand{\eeaa}{\end{eqnarray*}}
\newcommand{\bnad}{\begin{nad}}
\newcommand{\enad}{\end{nad}}

\usepackage{microtype}      

\usepackage[dvipsnames]{xcolor}

\title{Sub-Seasonal Climate Forecasting via Machine Learning: Challenges, Analysis, and Advances}
\date{}
\author[1]{Sijie~He \thanks{Equal Contribution}}
\author[1]{Xinyan~Li \textsuperscript{*}}
\author[2]{Timothy~DelSole}
\author[3]{Pradeep~Ravikumar}
\author[1]{Arindam~Banerjee}
\affil[1]{Department of Computer Science \& Engineering\\ 
  University of Minnesota, Twin Cities}
\affil[2]{Department of Atmospheric, Oceanic, and Earth Science\\
  George Mason University}
\affil[3]{Machine Learning Department\\
   Carnegie Mellon University}
\affil[ ]{Email: \texttt{\{hexxx893@umn.edu, lixx1166@umn.edu, tdelsole@gmu.edu,\protect\\ pradeepr@cs.cmu.edu, banerjee@cs.umn.edu\}}}

\begin{document}

\maketitle

\begin{abstract}
Sub-seasonal climate forecasting (SSF) focuses on predicting key climate variables such as temperature and precipitation in the 2-week to 2-month time scales. Skillful SSF would have immense societal value, in areas such as agricultural productivity, water resource management, transportation and aviation systems, and emergency planning for extreme weather events. However, SSF is considered more challenging than either weather prediction or even seasonal prediction. In this paper, we carefully study a variety of machine learning (ML) approaches for SSF over the US mainland. While atmosphere-land-ocean couplings and the limited amount of good quality data makes it hard to apply black-box ML naively, we show that with carefully constructed feature representations, even linear regression models, e.g., Lasso, can be made to perform well. Among a broad suite of 10 ML approaches considered, gradient boosting performs the best, and deep learning (DL) methods show some promise with careful architecture choices. Overall, suitable ML methods are able to outperform 
 the climatological baseline, i.e., predictions based on the 30-year average at a given location and time. Further, based on studying feature importance, ocean (especially indices based on climatic oscillations such as El Ni\~no) and land (soil moisture) covariates are found to be predictive, whereas atmospheric covariates are not considered helpful. 
\end{abstract}

\section{Introduction}

\begin{figure}[t]
\centering
\subfigure[Sources of Predictability]{\includegraphics[width=0.3\textwidth]{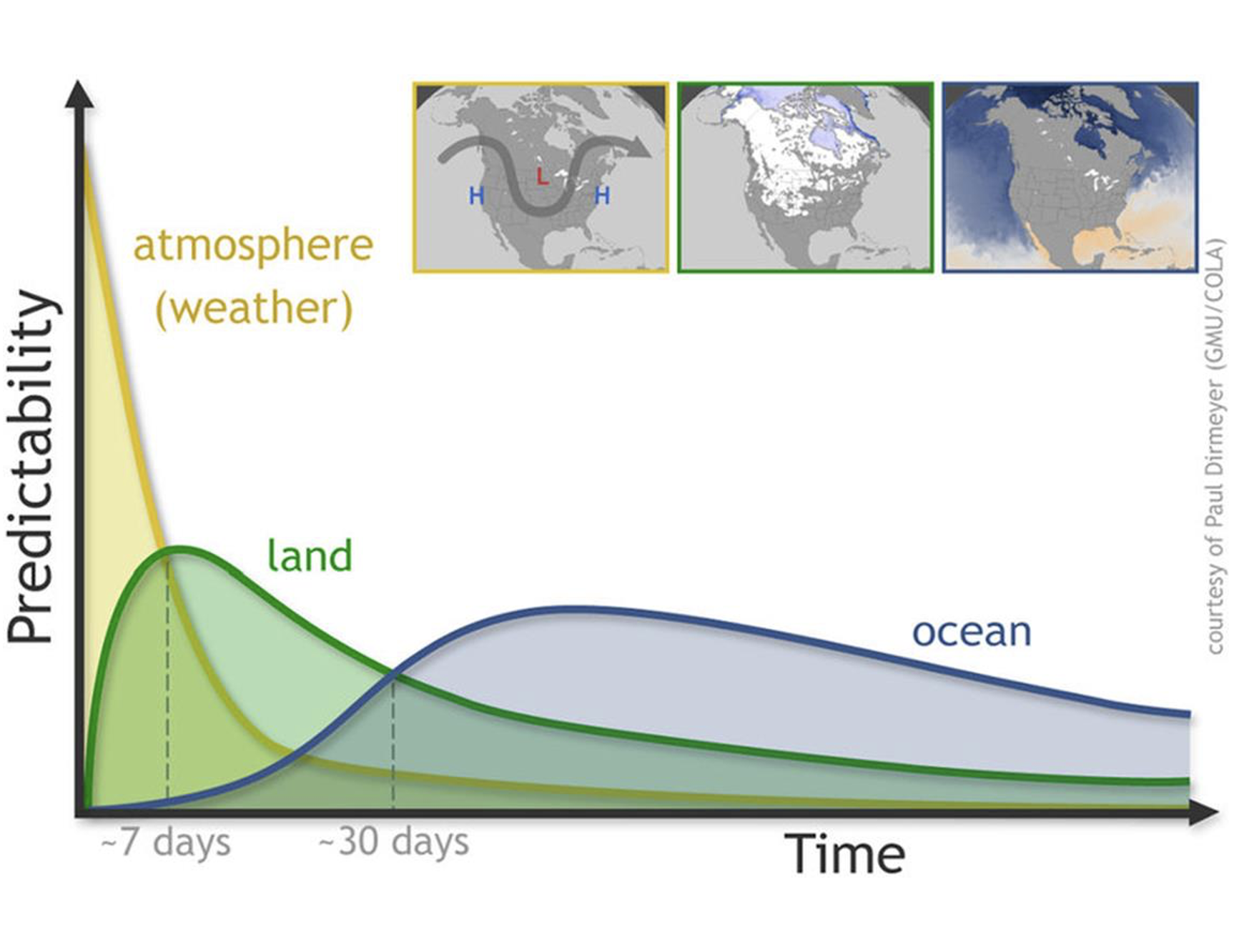}}
   \subfigure[MIC]{\includegraphics[width=0.3\columnwidth]{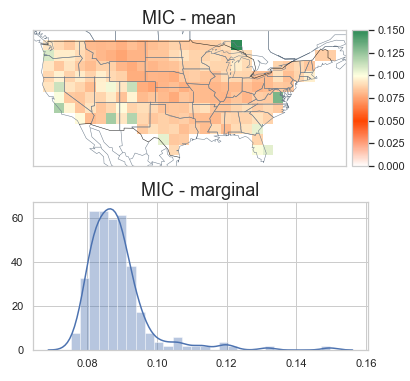}}
    \subfigure[Results of FNN and CNN]{\includegraphics[width=0.285\columnwidth]{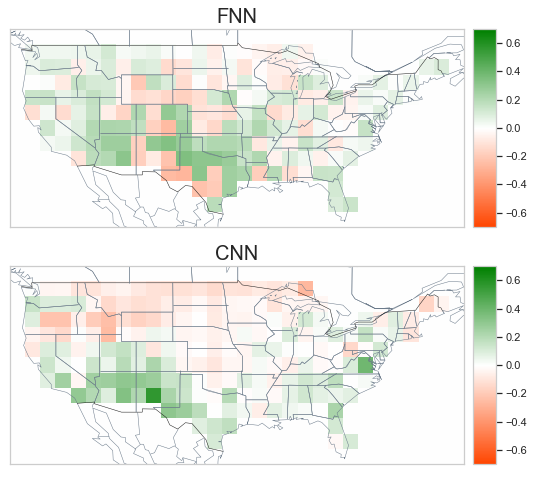}}
    \vspace{-2mm}
\caption{(a) Sources of predictability at different forecast time scales. Atmosphere is most predictive at weather time scales, whereas for SSF, land and ocean are considered important sources of predictability~\cite{predictability_s2s}. (b) Maximum information coefficient (MIC) \cite{reshef2011detecting} between temperature of week 3 \& 4 and week -2 \& -1. Small MICs ($\le 0.1$) at a majority of locations indicate little information shared between the most recent date and the forecasting target. (c) Predictive skills of Fully connected Neural Networks (FNN) and Convolutional Neural Networks (CNN), in terms of 
temporal cosine similarity (see definition in Section \ref{sec:exp_setup}), for temperature prediction over 2017-2018. Positive values closer to 1 (green) indicate better predictive skills.}
\label{fig:corr_tmp2m}
\end{figure}

Over the past few decades, major advances have been made in weather forecasts on time scales of days to about a week \cite{lorenc1986analysis,simmons2002some,nati10,boar16}. Similarly, major advances have been made in seasonal forecasts on time scales of 2-9 months~\cite{barnston2012skill}. However, making high quality forecasts of key climate variables such as temperature and precipitation on sub-seasonal time scales, defined here as the time range between 2-8 
weeks, has long been a gap in operational forecasting \cite{boar16}.
Skillful climate forecasts at sub-seasonal time scales would be of immense societal value, and would have an impact in a wide variety of domains including agricultural productivity, hydrology and water resource management, and emergency planning for extreme climate, etc. \cite{pomeroy2002prediction,klemm2017development}. The importance of sub-seasonal climate forecasting (SSF) has been discussed in great detail in two recent high profile reports from the National Academy of Sciences (NAS)~\cite{nati10,boar16}. Despite the scientific, societal, and financial importance of SSF, the progress on the problem has been restricted \cite{braman2013climate,de2014climate} partly because it has attracted less attention compared to weather and seasonal climate prediction. Also, SSF is arguably more difficult compared to weather or seasonal forecasting due to limited predictive information from land and ocean, and virtually no predictive information from the atmosphere, which forms the basis of numerical weather prediction (NWP) \cite{predictability_s2s,simmholl2002} (Figure \ref{fig:corr_tmp2m}(a)).

There exists great potential to advance sub-seasonal prediction using machine learning techniques, which has revolutionized statistical prediction in many other fields. Due in large part to this potential promise, a recently concluded real-time forecasting competition called the Sub-Seasonal Climate Forecast Rodeo, was sponsored by the Bureau of Reclamation in partnership with NOAA, USGS, and the U.S. Army Corps of Engineers \cite{hwang2019improving}. However, a direct application of standard black-box machine learning approaches to SSF can run into challenges due to the high-dimensionality and strong spatial correlation of the raw data from atmosphere, ocean, and land, e.g., Figure~\ref{fig:corr_tmp2m} shows that popular approaches such as Fully connected Neural Networks (FNN) and Convolutional Neural Networks (CNN) do not perform so well when directly applied to the raw data. Besides, sub-seasonal forecasting does not lie in the big data regime: about 40 years of reliable data exists for all climate variables, with each day corresponding to one data point, which totals less than 20,000 data points. Furthermore, different seasons have different predictive relations, and many climate variables have strong temporal correlations on daily time scales, further reducing the effective data size. Therefore, it is worth carefully and systematically investigating the capability of both classical and modern Machine Learning (ML) approaches including Deep Learning (DL) while keeping in mind the high-dimensionality, spatial-temporal correlations, and limited observational data available for SSF. While such a study can be extended to consider climate model output data at sub-seasonal time scales, we do not consider using model output data in the current study.

In this paper, we perform a comprehensive empirical study on ML approaches for SSF and discuss the challenges and advancements. Our main contributions are as follows:
\begin{itemize}
    \item We illustrate the difficulty of SSF due to the complex physical couplings as well as the unique nature of climate data, i.e., strong spatial-temporal correlation and high-dimensionality.

    \item We show that suitable ML models, e.g., XGBoost, to some extent, capture predictability for sub-seasonal time scales from climate data, and persistently outperform existing approaches in climate science, such as climatology and the damped persistence model.

    \item We demonstrate that even though DL models are not the obvious winner, they still show promising results with demonstrated improvements from careful architectural choices. With further improvements, DL models present a great potential topic for future research.
    \item We find that ML models tend to select covariates from the land and ocean, such as soil moisture and El Ni\~no indices, and rarely select atmospheric covariates, such as 500mb geopotential height or other indicators of the atmospheric general circulation.   

\end{itemize}


{\bf Organization of the paper.} We start with a review of related work in Section \ref{sec:related_work}. Section \ref{sec:ssf_difficulty} provides a formal description of the specific SSF problem targeted in this paper and 
demonstrates the difficulty of sub-seasonal climate forecasting for ML techniques. Next, we briefly discuss ML approaches we plan to investigate (Section~\ref{sec:models}) followed by details on data and experimental setup (Section~\ref{sec:exp_setup}). Subsequently, Section \ref{sec:results} presents experimental results, comparing the predictive skills over 10 ML models, including several DL models. Finally, we conclude in Section~\ref{sec:conclusion}.

\section{Related Work}

Although statistical models were used for weather prediction before the 1970s~\cite{Nebeker1995}, since the 1980s weather forecasting has been carried out using mainly physics-based dynamic system models~\cite{barnston2012skill}. More recently, there is a surge of application for ML approaches to both short-term weather forecasting~\cite{cofino2002bayesian,grover2015deep,radhika2009atmospheric}, and longer-term climate prediction~\cite{badr2014application,cohen2019s2s}. However, little attention has been paid on forecasting with sub-seasonal time scale which has been considered a ``predictability desert''~\cite{vitart2012}. Recently, ML techniques have made great strides in statistical prediction in many fields, so it is natural to investigate whether it can advance sub-seasonal climate prediction. In particular, many advances have occurred in high-dimensional sparse models and their variants which could be suitable for spatial-temporal climate data \cite{gonccalves2014multi,gonccalves2016multi,gonccalves2017spatial,delsole2017statistical,he2019interpretable}. Such models have been successfully applied to certain problems, e.g., predicting land temperature using oceanic climate data~\cite{delsole2017statistical,he2019interpretable}. Recently, promising progresses~\cite{hwang2019improving,he2019interpretable} have been seen on applying ML algorithms to solve SSF.

Since SSF can be formulated as a sequential modeling problem~\cite{sutskever2014sequence,venugopalan2015sequence}, 
bringing the core strength of DL-based sequential modeling, a thriving research area, has the great potential for a transformation in climate forecasting~\cite{ham2019deep,reichstein2019deep, schneider2017earth}. In the past decade, recurrent neural network (RNN)~\cite{funahashi1993approximation}, and long short-term memory (LSTM) models~\cite{gers1999learning}, are two of the most popular sequential models and have been successfully applied in language modeling and other seq-to-seq tasks~\cite{sundermeyer2012lstm}. Starting from~\cite{sutskever2014sequence,srivastava2015unsupervised}, the encoder-decoder structure with RNN or LSTM has become one of the most competitive algorithms for sequence transduction. The variants of such model that incorporate mechanisms like convolution~\cite{xingjian2015convolutional,shi2017deep} or attention mechanisms~\cite{attention} have achieved remarkable breakthroughs for audio synthesis, word-level language modeling, and machine translation~\cite{vaswani2017attention}. 

\label{sec:related_work}

\section{Sub-seasonal Climate Forecasting}
\label{sec:ssf_difficulty}
{\bf Problem statement.} In this paper, we focus on building temperature forecasting models at the forecast horizon of 15-28 days ahead, i.e., the average daily temperature of week 3 \& 4. The geographic region of interest is the US mainland (latitudes 25N-49N and longitudes 76W-133W) at a 2$^{\circ}$ by 2$^{\circ}$ resolution (197 grid points). For covariates, we consider climate variables, such as sea surface temperature, soil moisture, and geopotential height, etc., that can indicate the status of the three main components, i.e., land, ocean, and atmosphere. Table \ref{table: data} provides a detailed description. 

\begin{table}[t]
\caption{Description of climate variables and their data sources.}
\vspace{-3mm}
{\tiny
\begin{center}
\resizebox{\textwidth}{!}{
\begin{tabular}{c|c|c|c|c|c}
\hline
 Type &\makecell{Climate variable} &Description &Unit  & \makecell{Spatial coverage} & \makecell{Data Source} \\
\hline
\parbox[t]{2mm}{\multirow{13}{*}{\rotatebox[origin=c]{90}{Spatial-temporal}}}
&\makecell{tmp2m} &\makecell{Daily average\\ temperature at 2 meters}& C$^{\circ}$ &  \multirow{3}{*}{\makecell{US mainland}}&
\makecell{CPC Global Daily \\Temperature~\cite{fan2008global}}
\\
\hhline{~---~-}

&sm&\makecell{Monthly \\Soil moisture}  & mm & & \makecell{CPC Soil Moisture\\~\cite{soil_1,soil_2,soil_3}}\\
 \hhline{~-----}
 & sst & \makecell{Daily sea surface \\ temperature} &  C$^{\circ}$   & \makecell{North Pacific\\ \& Atlantic Ocean}    &  \makecell{Optimum Interpolation\\ SST (OISST)~\cite{reynolds2007daily}}\\
 \hhline{~-----}
& rhum& \makecell{Daily relative humidity\\near the surface\\(sigma level 0.995)} & $\%$& \multirow{5}{*}{\makecell{US mainland\\and North Pacific\\ \& Atlantic Ocean} }&\multirow{5}{*}{{\makecell{Atmospheric\\ Research \\Reanalysis\\ Dataset~\cite{reanalysis}}}}\\
\hhline{~---~~}
&\makecell{slp}& \makecell{Daily pressure\\ at sea level} & Pa &     &                  \\ 
\hhline{~---~~}
&\makecell{hgt10 \& hgt500} & \makecell{Daily geopotential height \\ at  10mb  and  500mb} & m &    & \\

\hline
\parbox[t]{2mm}{\multirow{6}{*}{\rotatebox[origin=c]{90}{Temporal}}} &MEI & \makecell{Bimonthly multivariate \\ENSO index}&\multirow{7}{*}{NA} &  \multirow{7}{*}{NA} &  \makecell{NOAA ESRL \\MEI.v2 ~\cite{zimmerman2016utilizing}}\\
\hhline{~--~~-}
& \makecell{Ni\~no 1+2, 3, \\3.4, 4}& \makecell{Weekly Oceanic \\Ni\~no Index (ONI)}&  & &  \makecell{NOAA National\\ Weather Service, CPC ~\cite{reynolds2007daily}}\\
\hhline{~--~~-}
& \makecell{NAO}& \makecell{Daily North Atlantic\\Oscillation index}&  & &  \makecell{NOAA National\\ Weather Service, CPC ~\cite{NAO_1,NAO_2}}\\
\hhline{~--~~-}
& \makecell{MJO phase \\ \& amplitude}& \makecell{Madden-Julian \\Oscillation index}&  & &  \makecell{Australian \\Government BoM \cite{MJO}}\\
\hline
\end{tabular}}
\end{center}}
\label{table: data}
\end{table}

\textbf{Difficulty of the problem.} To illustrate the challenge of SSF, we measure the dependence between the normalized average temperature of week -2 \& -1 (1-14 days in the past) and week 3 \& 4 (15-28 days in the ``future") at each grid by maximum information coefficient (MIC) \cite{reshef2011detecting}, an information theory-based measure of the linear or non-linear association between two variables. The values of MIC range between 0 and 1, and a small MIC value close to 0 indicates a weak dependence. To assess statistical significance, we apply moving block bootstrap~\cite{kunsch1989jackknife} to time series of 2-week average temperature at each grid point from 1986 to 2018, with the block size of 365 days. The top panel in Figure \ref{fig:corr_tmp2m}(b) illustrates the average MIC from 100 bootstrap over the US mainland, and the marginal distribution of all locations is shown at bottom. Small MIC values ($\le 0.1$), which indicates little predictive information shared between the most recent data and the forecasting target, to some extent, demonstrate how difficult SSF is.

From an ML perspective, applying black-box DL approaches naively to SSF is less likely to work due to the limited number of samples, and the high-dimensional and spatial-temporally correlated features. Figure \ref{fig:corr_tmp2m}(c) shows the performance of two vanilla DL models: Fully connected Neural Networks (FNN) with ReLU activation and Convolutional Neural Networks (CNN), in terms of the (temporal) cosine similarity between the prediction and the ground truth at each location over 2017-2018. For most locations, their cosine similarities are either negative or close to zero. In addition, we evaluate 10 ML models with suitable hyper-parameter tuning using another metric called relative $R^2$ (see formal definition in Appendix~\ref{sec:app_ssf_difficulty}), which compares the predictive skill of a model to the best constant prediction based on climatology, the 30 year average from historical training data. Most of the models do not get even positive relative $R^2$ (details are presented in Appendix~\ref{sec:app_ssf_difficulty}), indicating that they perform no better than the long term average. Such results are another good indication that accurate SSF is hard to achieve.


\section{Methods}

\textbf{Notation.} 
Let $t$ denote a date and $g$ denote a location. The target variable at time $t$ is denoted as $\y_t\in\R^G$, where $G$ represents the number of target locations. More specifically, $y_{g,t}$ is the normalized average temperature over time $t+15$ to $t+28$, i.e., weeks 3 \& 4 (details on normalization can be found in Section \ref{sec:exp_setup}). $X_{g,t}\in \R^p$ denotes the $p$-dimensional covariates designed for time $t$ and location $g$, which is also denoted as $X_t$ if the covariates are shared by all locations $g \in G$.

\textbf{ML (non-DL) models.} We compare the following ML (non-DL) models with DL models. 
\vspace{-2mm}

\begin{itemize}[leftmargin=*]
\setlength\itemsep{0em}
\item\textbf{MultiLLR \cite{hwang2019improving}.} MultiLLR introduces a multitask feature selection algorithm to remove the irrelevant predictors and integrates the remaining predictors linearly. For a location $g$ and target date $t^*$, its coefficient $\beta_g$ is estimated by $\hat{\beta}_g=\argmin_{\beta} \sum_{t\in \mathcal{D}} w_{t,g}(y_{g,t}-\beta^T X_{g,t})^2$, where $\mathcal{D}$ is the temporal span around the target date's day of the year and $w_{t,g}$ is the corresponding weight. In \cite{hwang2019improving}, an equal data point weighting ($w_{t,g}=1$) has been employed.

\begin{figure}[t]
    \centering
    \subfigure[Encoder (LSTM)-Decoder (FNN)]{\includegraphics[width=0.47\columnwidth]{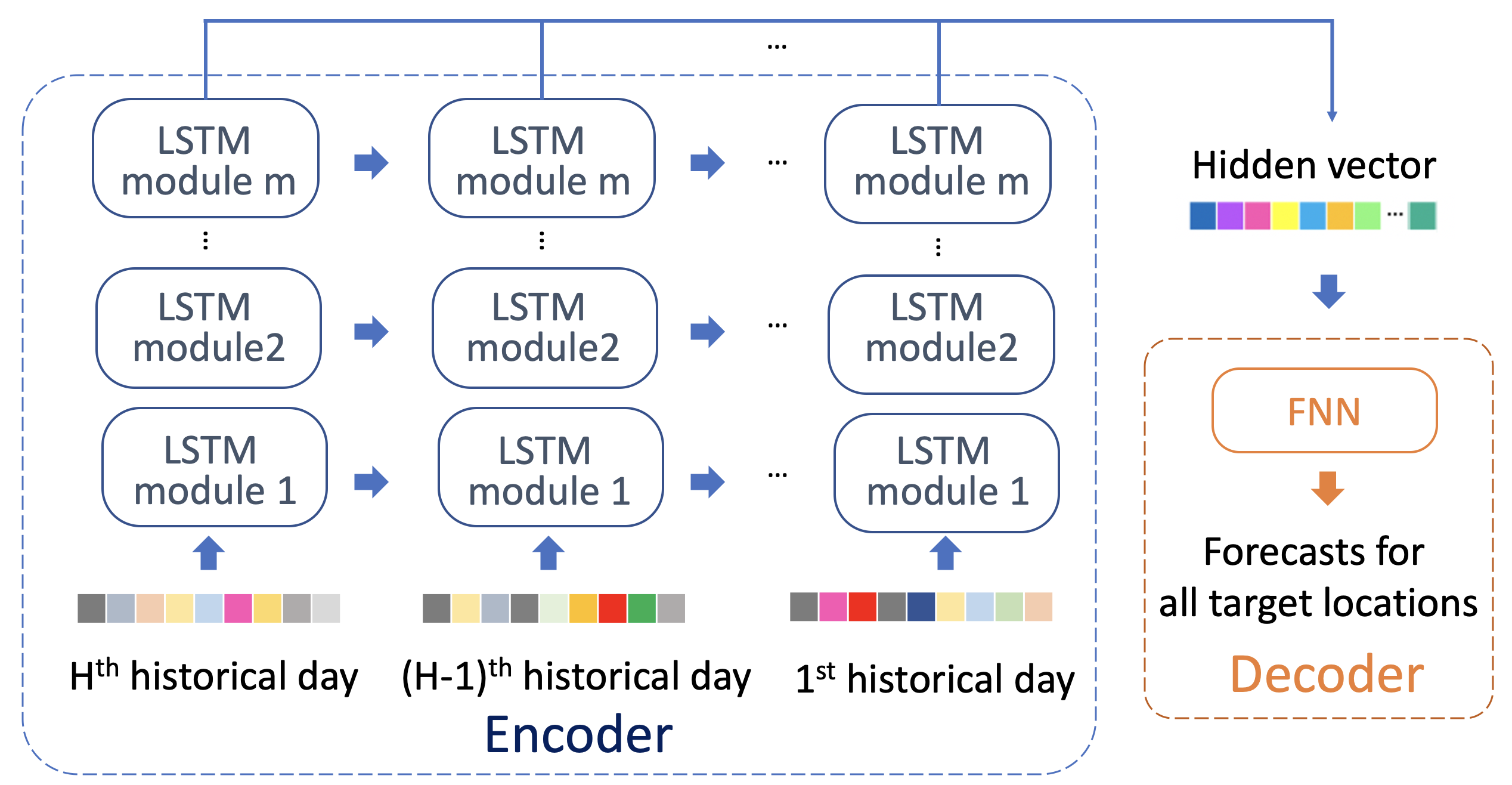}}
    \hspace{2mm}
    \subfigure[CNN-LSTM]{\includegraphics[width=0.47\columnwidth]{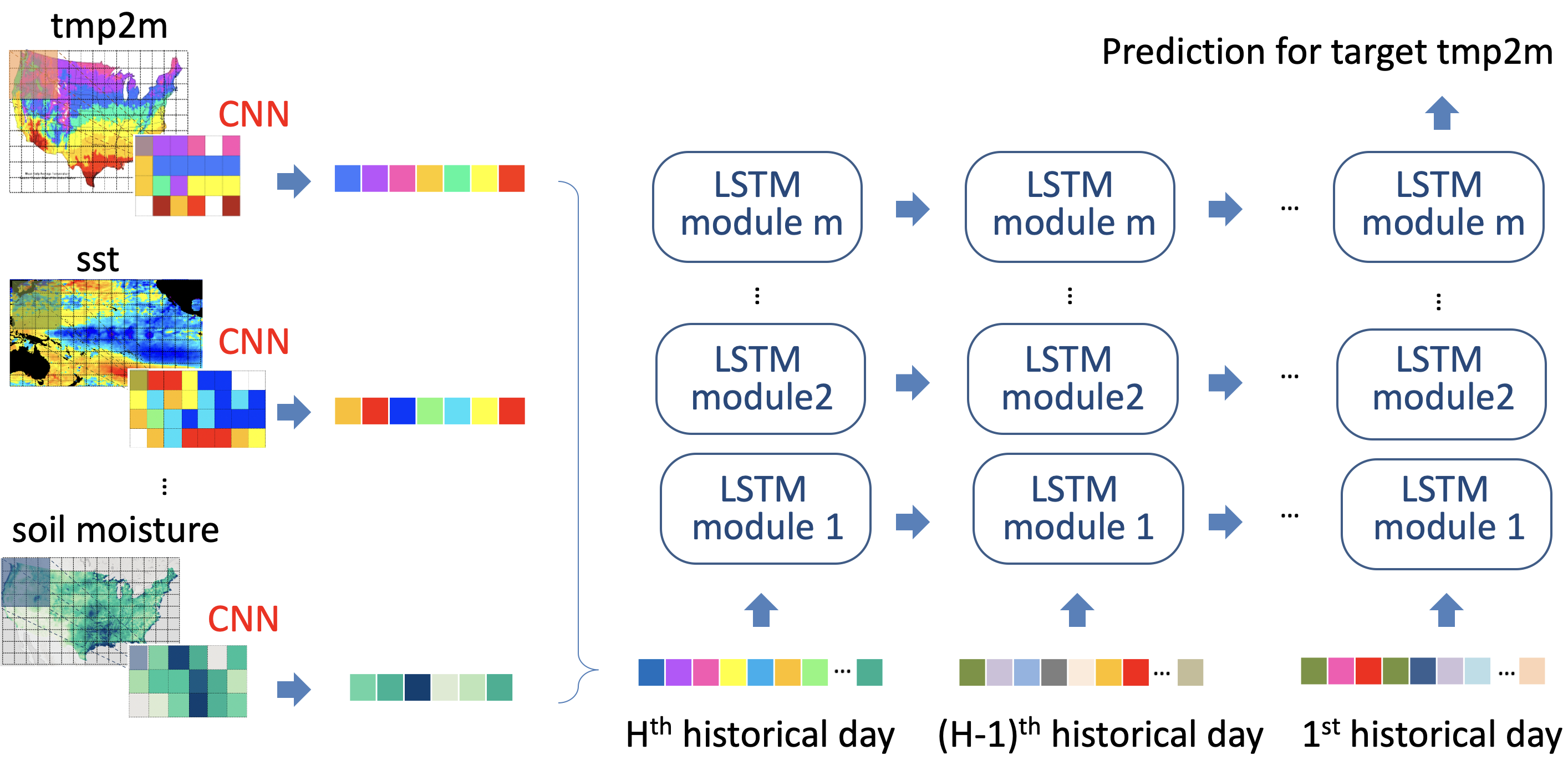}}
    \vspace{-2mm}
    \caption{Architectures of the designed DL models. (a) Encoder (LSTM)-Decoder (FNN) includes a few LSTM layers as the Encoder, and two fully connected layers as the Decoder. (b) CNN-LSTM consists of a few convolutional layers followed by an LSTM.
    }
     \label{fig:model_architecture}
\end{figure}

\item\textbf{AutoKNN \cite{hwang2019improving}.} An auto-regression model with weighted temporally local samples where the auto-regression lags are selected via a multitask k-nearest neighbor criterion. The method only takes historical measurements of the target variables as input. The nearest neighbors of each target date are selected based on an average of spatial cosine similarity computed over a history of $M=60$ days, starting one year prior to a target date $t^*$ (lag $l=365$). More precisely, the similarity between the target date $t^*$ and 
a date $t$ in the corresponding training set is formulated as 
$\text{sim}_t=\frac{1}{M}\sum_{m=0}^{M-1} \cos(\y_{t-l-m},\y_{t^*-l-m}),$ where $\cos(\y_{t_1},\y_{t_2})$ computes the (spatial) cosine similarity (see formal definition in Section \ref{sec:exp_setup}), evaluated over $G$ locations, between two given dates $t_1$ and $t_2$.

\item \textbf{Multitask Lasso \cite{lasso,jalali2013dirty}.} It assumes $\y_t = X_t\Theta^* + \epsilon$, where $\epsilon \in \R^G$ is a Gaussian noise vector and $\Theta^*\in \R^{p\times G}$ is the coefficient matrix for all locations. With $n$ samples, $\Theta^*$ is estimated by $\hat{\Theta}_n = \argmin_{\Theta\in \R^{p\times G}}\frac{1}{2n} \|Y - X\Theta\|_2^2 + \lambda_n \norm{\Theta}_{21}$ with $X\in \R^{n\times p}$ and $Y\in\R^{n\times G}$. $\lambda_n$ is a penalty parameter and the corresponding penalty term is computed as $||\Theta||_{21} = \sum_{i} (\sum_{j} \Theta_{ij}^2)^{1/2}$.

\item \textbf{Gradient Boosted Trees (XGBoost) \cite{friedman2001greedy,xgboost}.} A functional gradient boosting algorithm using regression tree as its weak learner. The algorithm starts with one weak learner and iteratively adds new weak learners to approximate functional gradients. The final ensemble model is constructed by a weighted summation of all weak learners. It is implemented using the Python package XGBoost. 

\item \textbf{State-of-the-art Climate Baseline.} We consider two baselines from climate science perspective, both are Least Square (LS) linear regression models \cite{weisberg2005applied}. The first model has predictors as climate indices, such as NAO index and Ni\~no indices, which are used to monitor ocean conditions. The predictor of the second model is the most recent anomaly of the target variable, i.e., anomaly temperature of week -2 \& -1, with which the model, also known as \textit{damped persistence} \cite{van2007empirical} in climate science, is essentially a first-order autoregressive model.

\end{itemize}
\vspace{-2mm}
\textbf{DL models.} As shown in Figure \ref{fig:corr_tmp2m}(c), it is hard for vanilla deep learning models like FNN and CNN to achieve high prediction accuracy. Therefore, we design two variants of DL models to improve the performance, namely Encoder (LSTM)-Decoder (FNN) and CNN-LSTM.
\vspace{-2mm}
\begin{itemize}[leftmargin=*]
\setlength\itemsep{0em}
\item\textbf{Encoder (LSTM)-Decoder (FNN).} Inspired by Autoencoder widely used in sequential modeling \cite{sutskever2014sequence}, we design the Encoder (LSTM)-Decoder (FNN) model, of which the architecture is shown in Figure \ref{fig:model_architecture}(a). Input of the model is features extracted spatially from covariates using unsupervised methods like Principal Component Analysis (PCA). The temporal components of covariates are handled by feeding features of each historical date into an LSTM Encoder recurrently. Then, the output of each date from LSTM is sent jointly to a two-layer FNN network using ReLU as an activation function. The output of the FNN Decoder is the predicted average temperature of week 3 \& 4 over all target locations. 

\item\textbf{CNN-LSTM.}  The proposed CNN-LSTM model directly learns the representations from the spatial-temporal data using CNN components \cite{lecun1998gradient}. Shown in Figure \ref{fig:model_architecture}(b), CNN extracts features for each climate variable at all historical dates separately. Then, the extracted features from the same date are collected and fed into an LSTM model recurrently. The temperature prediction for all target locations is done by an FNN layer taking the output of the LSTM's last layer from the latest input. 

\end{itemize}

\label{sec:models}

\section{Data and Experimental Setup}

\textbf{Data description.} Climate agencies across the world maintain multiple datasets with different formats and resolutions. Climate variables (Table~\ref{table: data}) have been collected from diverse data sources and converted into a consistent format. Temporal variables, e.g., Ni\~no indices, are interpolated to a daily resolution, and spatial-temporal variables are interpolated to a spatial resolution of  $0.5^{\circ}$ by $0.5^{\circ}$.

{\bf Preprocessing.} For spatial-temporal variables, we first extract the top 10 principal components (PCs) as features based on PC loadings from 1986 to 2016 (for details, refer to Appendix~\ref{sec:app_exp_setup}). Next, we 
normalize the data by z-scoring at each location and each date with the corresponding mean and standard deviation of the corresponding day of the year over 1986-2016 separately. Note that both training and test sets are z-scored using the mean and standard deviations of the same 30-year historical data. Temporal variables, e.g., Ni\~no indices, are directly used without normalization.

\textbf{Feature set construction.} We combine the PCs of spatial-temporal covariates with temporal covariates into a sequential feature set, which consists not only covariates of the target date, but also covariates of the $7^{th}$, $14^{th}$, and $28^{th}$ day previous from the target date, as well as the day of the year of the target date in the past 2 years and both the historical past and future dates around the day of the year of the target date in the past 2 years (see Appendix~\ref{sec:app_exp_setup} for a detailed example).

\textbf{Evaluation pipeline.} Predictive models are created independently for each month in 2017 and 2018. To mimic a live system, we generate 105 test dates during 2017-2018, one for each week, and group them into 24 test sets by their month of the year. Given a test set, our evaluation pipeline consists of two parts: (1) ``5-fold'' training-validation pairs for hyper-parameter tuning, based on a ``sliding-window'' strategy designed for time-series data. Each validation set uses the data from the same month of the year as the test set, and we create 5 such set from dates in the past 5 years. Their corresponding training sets contain 10 years of data before each validation set; (2) the training-test pair, where the training set, including 30-year data in the past, ends 28 days before the first date in the test set. We share more explanations, including a pictorial example, in Appendix~\ref{sec:app_exp_setup}.

\textbf{Evaluation metrics.} Forecasts are evaluated by cosine similarity, a widely used metric in weather forecasting, between $\hat{\y}$, a vector of predicted values, and $\y^*$, the corresponding ground truth. It is computed as $\frac{\langle \hat{\y},\y^*\rangle}{\norm{\hat{\y}}_2\norm{\y^*}_2}$, where $\langle \hat{\y},\y^* \rangle$ denotes the inner product between the two vectors. If $\hat{\y}$ represents the predicted values for a period of time at one location, it becomes \textit{temporal cosine similarity} which assesses the prediction skill at a specific location. Whereas, if $\hat{\y}$ contains the predicted values for all target locations at one date, it becomes \textit{spatial cosine similarity} measuring the prediction skill at that date. To get a better intuition, one can view spatial and temporal cosine similarity as spatial and temporal correlation respectively, measured between two centered vectors.
\label{sec:exp_setup}

\section{Experimental Results}

We compare the predictive skills of 10 ML models on SSF over the US mainland. In addition, we discuss a few aspects that impact the ML models the most, and the evolution of our DL models.

\vspace{-2mm}
\subsection{Results of all methods} 
\vspace{-2mm}
\textbf{Temporal results.} Table \ref{tab:cos_all} lists the mean, the median, the 0.25 quantile, the 0.75 quantile, and their corresponding standard errors of spatial cosine similarity of all methods. Additional results based on relative $R^2$ can be found in Appendix \ref{sec:app_results}. XGBoost, Encoder (LSTM)+Decoder (FNN) and Lasso accomplish higher predictive skills than other presented methods, and can outperform climatology and two climate baseline models, i.e., LS with NAO \& Ni\~no, and damped persistence. Overall, XGBoost achieves the highest predictive skill in terms of both the mean and the median, demonstrating its predictive power. Surprisingly, linear regression with a proper feature set has good predictive performance. Even though DL models are not the obvious winner, with careful architectural selections, they still show promising results.

\begin{table}[t]
  \caption{Comparison of spatial cosine similarity of tmp2m forecasting for test sets over 2017-2018. XGBoost and Encoder (LSTM)-Decoder (FNN) have the best performance. Models achieve better performance using temporally global set compared to temporally local set. 
 }
  \resizebox{\textwidth}{!}{
  \begin{tabular}{c|c c c c}
  \hline
    Model & Mean(se) & Median (se) & 0.25 quantile (se) & 0.75 quantile (se)\\
    \hline
 \multicolumn{5}{c}{\textbf{Temporally Global Dataset}}\\
    \hline
    \textbf{XGBoost - one day}&\textbf{ 0.3044(0.03)}& \textbf{0.3447(0.05)}&  \textbf{0.0252(0.05)}&\textbf{ 0.5905(0.04})\\
  Lasso - one day &  0.2499(0.04)& 0.2554(0.06) & -0.0224(0.05) &0.5604(0.06)\\
    \hline
     \textbf{Encoder (LSTM)-Decoder (FNN)} &\textbf{ 0.2616 (0.04}) & \textbf{0.2995 (0.07)}
&\textbf{-0.0719 (0.06)} & \textbf{0.6310 (0.05)}\\

     FNN & 0.0792(0.01) & 0.0920(0.02) & 0.0085(0.02) &0.1655(0.02) \\
     CNN &0.1688(0.04) &0.2324(0.06)& -0.0662(0.06)& 0.4768(0.04)\\
     CNN-LSTM &  0.1743(0.04) &0.2867(0.06) & -0.1225(0.07) &0.5148(0.04)\\
  
    \hline
\textbf{LS with NAO \& Ni\~no }&\textbf{ 0.2415(0.03)} & \textbf{0.3169(0.04)}& \textbf{0.0454(0.05)} &\textbf{0.4624(0.03)}\\
 Damped persistence &0.2009(0.04)  &0.2310(0.06)  & -0.0884(0.06) & 0.5335(0.05))\\
    \hline 
    MultiLLR &0.0684 (0.03) & 0.1046 (0.05)
&-0.1764 (0.06) & 0.3156 (0.04)\\
AutoKNN &0.1457 (0.03) & 0.1744 (0.05)
&-0.1018 (0.06) & 0.4000 (0.04)\\
    \hline
\multicolumn{5}{c}{\textbf{Temporally Local Dataset}}\\
\hline
XGBoost - one day &0.1965(0.04)& 0.2345(0.05)&  -0.0636(0.06)& 0.5178(0.05)\\
 Lasso - one day & 0.1631(0.04) &0.2087(0.06) & -0.1178(0.05)& 0.5059(0.05)\\ 
Encoder (LSTM)-Decoder (FNN) & 0.1277 (0.04) & 0.1272 (0.06)
&-0.1558 (0.06) & 0.4971 (0.06)\\
 \hline
\end{tabular}}
\label{tab:cos_all}
\end{table}

\begin{figure}[t]
\centering
\includegraphics[width=\textwidth]{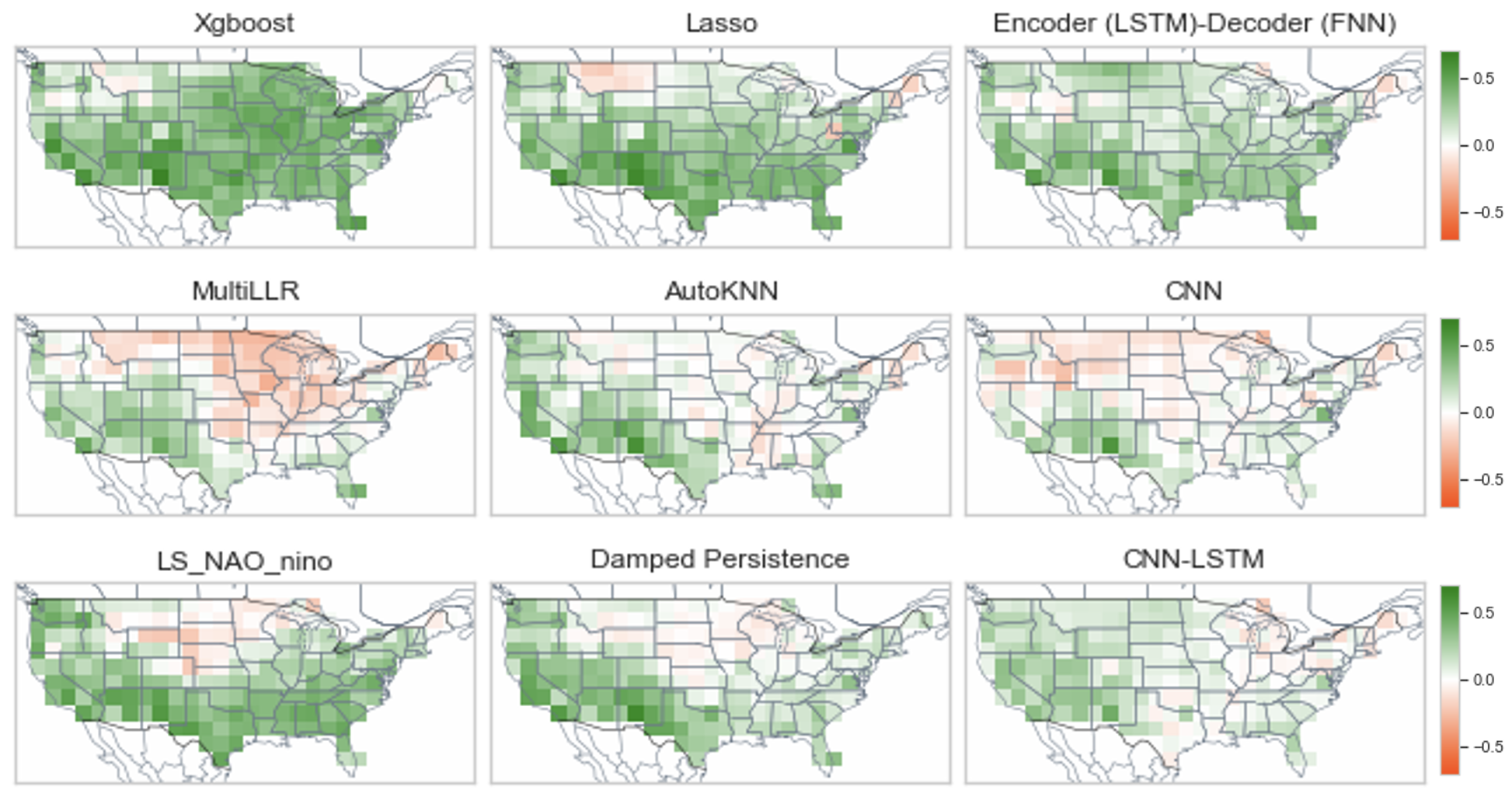}
\vspace{-5mm}
\caption{Temporal cosine similarity over the US mainland of ML models discussed in Section \ref{sec:models} for temperature prediction over 2017-2018. Large positive values (green) closer to 1 indicates better predictive skills. Overall, XGBoost and Encoder (LSTM)-Decoder (FNN) perform the best. Qualitatively, coastal and south regions are easier to predict than inland regions (e.g., Midwest).}
\label{fig:spatial_result_all}
\end{figure}

\textbf{Spatial results.} Figure~\ref{fig:spatial_result_all} shows the temporal cosine similarity of all methods evaluated on test sets described in Section \ref{sec:exp_setup}. Among all methods, XGBoost and the Encoder (LSTM)-Decoder (FNN) achieve the overall best performance, regarding the number of locations with positive temporal cosine similarity. Qualitatively, coastal and south regions, in general, are easier to predict compared to inland regions (e.g., Midwest). Such a phenomenon might be explained by the influence of the slow-moving component, i.e., Pacific and Atlantic Ocean. Such component exhibits inertia or memory, in which anomalous condition can take relatively long period of time to decay. However, each model has its own favorable and disadvantageous regions. For example, XGBoost and Lasso do poorly in Montana, Wyoming, and Idaho, while Encoder (LSTM)-Decoder (FNN) performs much better on those regions. The observations naturally imply that the ensemble of multiple models is a promising future direction.

\textbf{Comparison with the state-of-the-art methods.} MultiLLR and AutoKNN are two state-of-the-art methods designed for SSF on western US \cite{hwang2019improving}. Both methods have shown good forecasting performance on the original target region. However, over the inland region (Midwest), Northeast, and South region, the methods do not perform so well (Figure~\ref{fig:spatial_result_all}). To be fair, even though a similar set of climate variables have been used in our work compared to the original paper \cite{hwang2019improving}, how we prepossess the data and construct the feature set are slightly different. Such differences may lead to relatively poor performance for these two methods, especially for MultiLLR.

\subsection{Analysis and exploration}

We analyze and explore several important aspects that could influence the performance of ML models.

\textbf{Temporally ``local'' vs. ``global'' dataset.} Our training set consists of 
all calendar months over the past 30 years, which we refer to as the temporally ``global'' dataset. Another way to construct the training set is to consider calendar months only within the temporal neighborhood of the test date. For instance, to build a predictive model to forecast June in 2017, the training set can only contains dates in June (from earlier years), and months that are close to June, e.g., April, May, July, and August, over the past 30 years. Such a construction account for the seasonal dependence of predictive relations, for example summer predictions are not trained with winter data. We name such dataset as a temporally ``local'' dataset. A comparison between the ``global'' and ``local'' datasets has been listed in Table \ref{tab:cos_all} where a significant drop in cosine similarity can be noticed when using ``local'' dataset for all of our best predictors, including XGBoost, Lasso, and Encoder (LSTM)-Decoder (FNN). We suspect such performance drop from ``global'' to ``local'' dataset may come from the reduction in the number of effective samples.

\begin{figure}[t]
    \centering
    \subfigure[XGBoost]{\includegraphics[width=0.45\columnwidth]{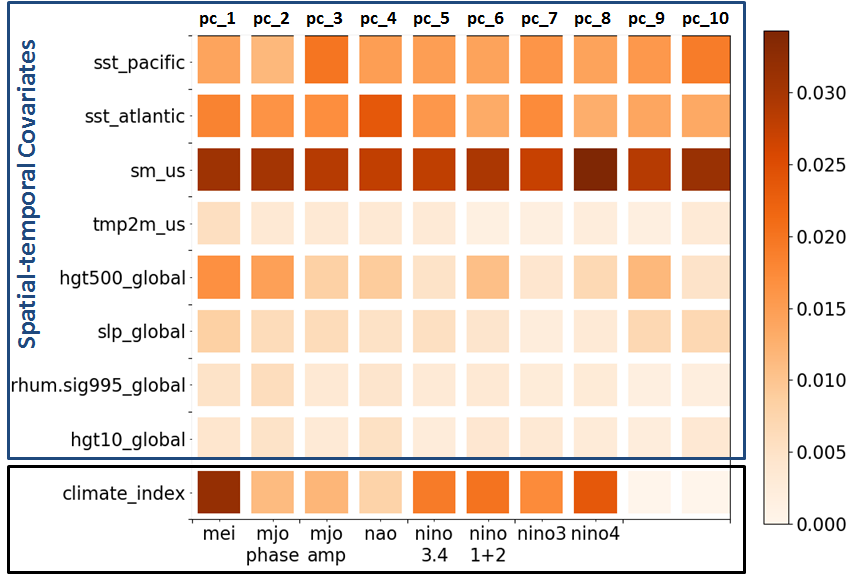}}
    \hspace{3mm}
    \subfigure[Lasso]{\includegraphics[width=0.43\columnwidth]{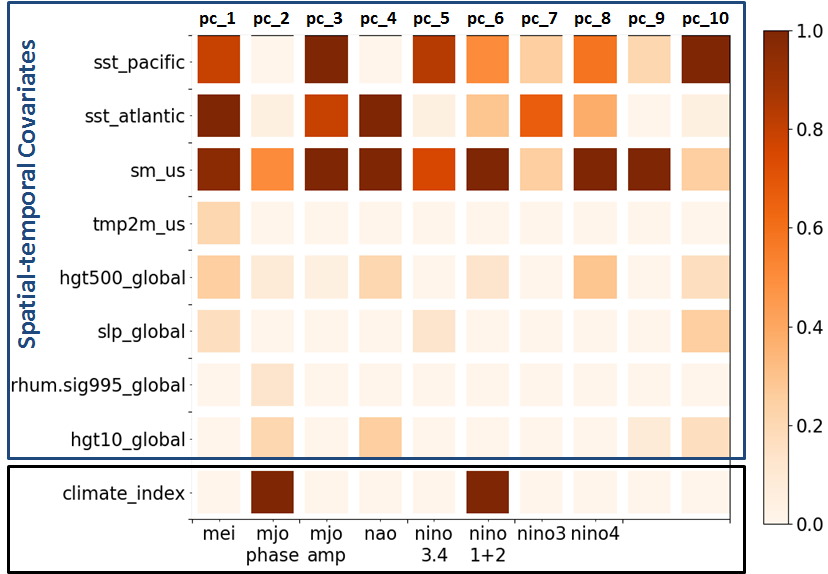}}
    \vspace{-2mm}
    \caption{Feature importance scores computed from (a) XGBoost and (b) Lasso. Darker color means a covariate is of the higher importance. The first 8 rows contains the top 10 principal components (PCs) extracted from 8 spatial-temporal covariates respectively, and the last row includes all the temporal indices. Land component, e.g., soil moisture ($3^{rd}$ row from the top) and ocean components, e.g., sst (Pacific and Atlantic) and some climate indices are the most commonly selected covariates.}
     \label{fig:feature_importance}
\end{figure}

\begin{figure}[h]
    \centering
    \subfigure[XGBoost]{\includegraphics[width=0.46\columnwidth]{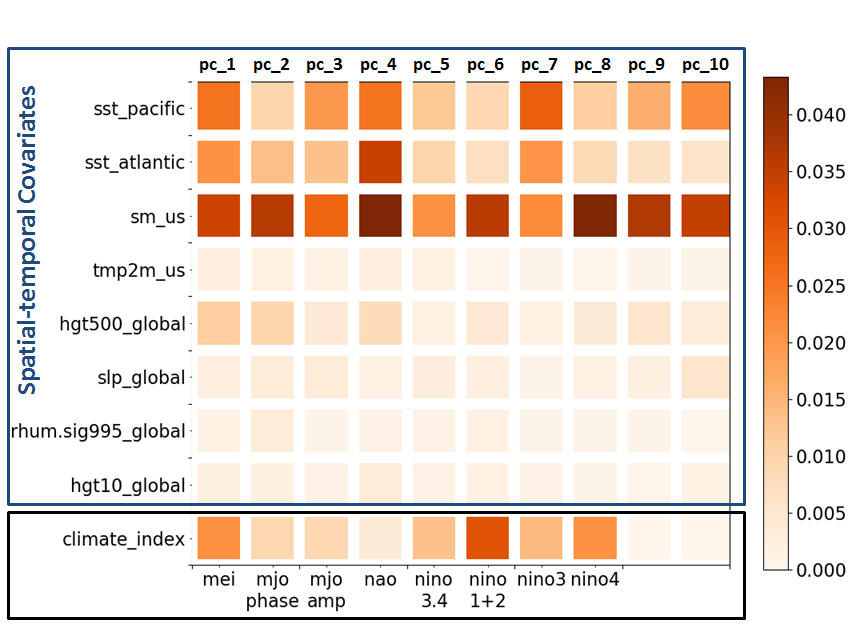}}
    \hspace{3mm}
    \subfigure[Lasso]{\includegraphics[width=0.45\columnwidth]{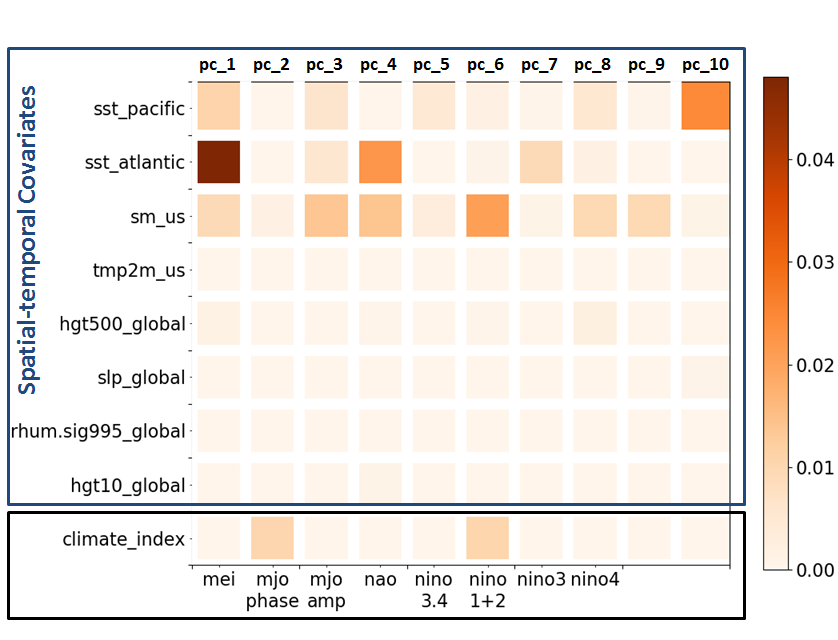}}
    \vspace{-2mm}
    \caption{Shapley values computed from (a) XGBoost and (b) Lasso. Darker color means a covariate is of the higher importance. The observation that the most commonly selected covariates are from land and ocean are consistent with Figure \ref{fig:feature_importance}.}
    \label{fig:shapley_values}
\end{figure}

\textbf{Feature importance.} At sub-seasonal time scales, 
it is widely believed \cite{predictability_s2s,delsole2017predictability} that land and ocean are the important sources of predictability, while the impact of atmosphere is limited (Figure~\ref{fig:corr_tmp2m} (a)). We study which covariate(s) are important, considered by ML models, based on the feature importance score. In particular, we compute the feature importance score from 2 ML models, XGBoost and Lasso (Figure~\ref{fig:feature_importance}). For XGBoost, the importance score is computed using the average information gain across all tree nodes a feature/covariate splits, while for Lasso, we simply count the non-zero coefficients of each model. The reported feature importance score is the average over 24 models (one per month in 2017-2018). In addition, we also provide measurement of feature importance based on Shapley value  (Figure~\ref{fig:shapley_values}), a concept from game theory~\cite{lipovetsky2001shapley}. To determine the Shapley value of a given feature, we compute the prediction difference between a model trained with and without that feature. Since the effect of suppressing a feature also depends on other features, we have to consider all possible subsets of other features, and compute the Shapley values as a weighted average of all possible differences. As shown in Figure \ref{fig:feature_importance} and \ref{fig:shapley_values}, among all covariates, soil moisture ($3^{rd}$ row from the top) is the variable that has been constantly selected by both models. Another set of important covariates is the family of Ni\~no indices. A LS model using those indices alone as predictors performs fairly well (Table \ref{tab:cos_all}). Sea surface temperatures (of Pacific and Atlantic) are also commonly selected. Such observations indicate that ML models pick up ocean-based covariates, some land-based covariates, and almost entirely ignore the atmosphere-related covariates, which are well aligned with domain knowledge ~\cite{predictability_s2s,delsole2017predictability}.

\textbf{The influence of feature sequence length.} To adapt the usage of LSTM, we construct a sequential feature set, which consists not only the target date, but also 17 other dates preceding the target date. However, other ML models, e.g., XGBoost and Lasso, which are not designed to handle sequential data, experience a drastic performance drop when we include more information from the past. More precisely, by including covariates from the full historical sequence, the performance of XGBoost drops approximately 50\% compared to the XGBoost model using covariates from the most recent date only. A possible explanation for such performance degradation, as we increase the feature sequence length, is that both models weight covariates from different dates exactly the same without considering temporal information, thus some irrelevant historical information might mislead the model. In Appendix~\ref{sec:app_results}, we provide a more detailed comparison among results obtained from various sequence lengths.

\subsection{What happened with DL models?}

As discussed in Section \ref{sec:ssf_difficulty}, applying black-box DL models naively does not work well for SSF. The improvement (Table \ref{tab:cos_all}), as we evolve from FNN to CNN-LSTM, and finally to Encoder (LSTM)-Decoder (FNN), demonstrates how the network architecture plays an important role, and leaves us plenty of space for further advancement. Below we focus on the discussion of feature representation, and the architecture design for sequence modeling. More discussion is included in Appendix~\ref{sec:app_results}.

\textbf{Feature representation: CNN vs. PCA.} Since SSF can be considered as a spatial-temporal prediction problem, CNN~\cite{lecun1998gradient} is a natural choice to handle the spatial aspect of each climate covariate by viewing it as a map, and can be applied as a ``supervised'' way for learning feature representation. However, our results imply that models with CNN has limited predictive skills regarding both spatial and temporal cosine similarity. CNN, while doing convolution using a small kernel, mainly focus on spatially localized regions. However, the strong spatial correlation of climate variables restricts the effectiveness of CNN kernels on feature extraction. Meanwhile, PCA, termed Empirical Orthogonal Functions (EOF)~\cite{lorenz1956empirical,von2001statistical} in climate science, is a commonly used ``unsupervised'' feature representation method, which focuses on low-rank modeling of spatial covariance structure revealing spatial connection. Our results (Table \ref{tab:cos_all}) illustrate that PCA-based models have higher predictive skills than CNN-based models, verifying that PCA is a better technique for feature extraction in SSF.

\textbf{Sequential modeling: Encoder-Decoder.} With features extracted by PCA,  we formulate SSF as a sequential modeling problem \cite{sutskever2014sequence}, where the input is the covariates sequence described in Section \ref{sec:exp_setup}, and the output is the target variable. Due to its immense success in sequential modeling \cite{srivastava2015unsupervised,venugopalan2015sequence}, the standard Encoder-Decoder, where both Encoder and Decoder are LSTM \cite{LSTM}, is the first architecture we investigate. Unfortunately, the model does not perform well and suffers from over-fitting, possibly caused by overly complex architecture. To reduce the model complexity, we replace the LSTM Decoder with an FNN Decoder which takes only the last step of the output sequence from the Encoder. Such change leads to an immediate boost of predictive performance. However, the input of the FNN Decoder mainly contains information encoded from the latest day in the input sequence and can only embed limited amount of historical information owing to the recurrent architecture of LSTM.  To further improve the performance, we adjust the connection between Encoder and Decoder, such that FNN Decoder takes every step of the output sequence from LSTM Encoder, which makes a better use of historical information. Eventually, this architecture achieves the best performance among all investigated Encoder-Decoder variants (see a detailed comparisons in Appendix~\ref{sec:app_results}).

\label{sec:results}

\section{Conclusion}
In this paper, we investigate the great potential to advance sub-seasonal climate forecasting using ML techniques. SSF, the skillful forecasts of temperature on the time range between 2-8 weeks, is a challenging task, since it is beyond the limit of atmospheric predictability commonly used in short-term weather forecasting. Besides, SSF is typically a high-dimensional problem with limited effective samples due to strong spatial-temporal correlation within climate data. We conduct a comprehensive analysis of 10 different ML models, including a few DL models. Empirical results show the gradient boosting model XGBoost, the DL model Encoder (LSTM)-Decoder (FNN), and linear models, such as Lasso, consistently outperform state-of-the-art forecasts. XGBoost has the highest skill over all models, and demonstrates its predictive power. ML models are capable of picking the climate variables from important sources of predictability in SSF, identified by climate scientists. In addition, DL models, with demonstrated improvements from careful architectural choices, are great potentials for future research.

\label{sec:conclusion}


\section{Broader Impact}

Skillful (i.e., accurate) climate forecasts on sub-seasonal time scales would have immense societal value. For instance, sub-seasonal forecasts of temperature and precipitation could be used to assist farmers in determining planting dates, irrigation needs, expected market conditions, anticipating pests and disease, and assessing the need for insurance. Emergency and disaster-relief supplies can take weeks or months to pre-stage, so skillful forecasts of areas that are likely to experience extreme weather a few weeks in advance could save lives. More generally, skillful sub-seasonal forecasts also would have beneficial impacts on agricultural productivity, hydrology and water resource management, transportation and aviation systems, emergency planning for extreme climate such as Atlantic hurricanes and midwestern tornadoes, among others \cite{pomeroy2002prediction,klemm2017development}. Inaccurate spatial-temporal forecasts associated with extreme weather events and associated disaster relief planning can be expensive both in terms of loss of human lives as well as financial impact. On a more steady state basis, water resource management and planning agricultural activities can be made considerably more precise and cost effective with skillful sub-seasonal climate forecasts.

\section*{Acknowledgement}
The research was supported by NSF grants OAC-1934634, IIS-1908104, IIS-1563950, IIS-1447566, IIS-1447574, IIS-1422557, CCF-1451986. The authors would like to acknowledge the computing support from the Minnesota Supercomputing Institute (MSI) at the University of Minnesota. 

\bibliographystyle{plain}
\bibliography{reference}

\begin{thebibliography}{10}

\bibitem{badr2014application}
Hamada~S Badr, Benjamin~F Zaitchik, and Seth~D Guikema.
\newblock Application of statistical models to the prediction of seasonal
  rainfall anomalies over the sahel.
\newblock {\em Journal of Applied meteorology and climatology}, 53(3):614--636,
  2014.

\bibitem{attention}
Dzmitry Bahdanau, Kyunghyun Cho, and Yoshua Bengio.
\newblock Neural machine translation by jointly learning to align and
  translate.
\newblock In {\em Conference Track Proceedings of the 3rd International
  Conference on Learning Representations (ICLR)}, 2015.

\bibitem{NAO_1}
Anthony~G Barnston and Robert~E Livezey.
\newblock Classification, seasonality and persistence of low-frequency
  atmospheric circulation patterns.
\newblock {\em Monthly weather review}, 115(6):1083--1126, 1987.

\bibitem{barnston2012skill}
Anthony~G Barnston, Michael~K Tippett, Michelle~L L'Heureux, Shuhua Li, and
  David~G DeWitt.
\newblock Skill of real-time seasonal enso model predictions during 2002--11:
  Is our capability increasing?
\newblock {\em Bulletin of the American Meteorological Society},
  93(5):631--651, 2012.

\bibitem{braman2013climate}
Lisette~Martine Braman, Maarten~Krispijn van Aalst, Simon~J Mason, Pablo
  Suarez, Youcef Ait-Chellouche, and Arame Tall.
\newblock Climate forecasts in disaster management: Red cross flood operations
  in west africa, 2008.
\newblock {\em Disasters}, 37(1):144--164, 2013.

\bibitem{xgboost}
Tianqi Chen and Carlos Guestrin.
\newblock Xgboost: A scalable tree boosting system.
\newblock In {\em Proceedings of the 22nd ACM SIGKDD International Conference
  on Knowledge Discovery and Data Mining (SIGKDD)}, pages 785--794, 2016.

\bibitem{cofino2002bayesian}
Antonio~S. Cofıno, Rafael Cano, Carmen Sordo, and Jos{\'e}~M. Guti{\'e}rrez.
\newblock Bayesian networks for probabilistic weather prediction.
\newblock In {\em In Proceedings of the 15th Eureopean Conference on Artificial
  Intelligence (ECAI)}, pages 695--699, 2002.

\bibitem{cohen2019s2s}
Judah Cohen, Dim Coumou, Jessica Hwang, Lester Mackey, Paulo Orenstein, Sonja
  Totz, and Eli Tziperman.
\newblock S2s reboot: An argument for greater inclusion of machine learning in
  subseasonal to seasonal forecasts.
\newblock {\em Wiley Interdisciplinary Reviews: Climate Change}, 10(2):e00567,
  2019.

\bibitem{de2014climate}
Erin~Coughlan de~Perez and Simon~J Mason.
\newblock Climate information for humanitarian agencies: Some basic principles.
\newblock {\em Earth Perspectives}, 1(1):11, 2014.

\bibitem{delsole2017statistical}
Timothy DelSole and Arindam Banerjee.
\newblock Statistical seasonal prediction based on regularized regression.
\newblock {\em Journal of Climate}, 30(4):1345--1361, 2017.

\bibitem{delsole2017predictability}
Timothy Delsole and Michael Tippett.
\newblock Predictability in a changing climate.
\newblock {\em Climate Dynamics}, 10 2017.

\bibitem{soil_3}
Yun Fan and Huug van~den Dool.
\newblock Climate prediction center global monthly soil moisture data set at
  0.5 resolution for 1948 to present.
\newblock {\em Journal of Geophysical Research: Atmospheres}, 109(D10), 2004.

\bibitem{fan2008global}
Yun Fan and Huug Van~den Dool.
\newblock A global monthly land surface air temperature analysis for
  1948--present.
\newblock {\em Journal of Geophysical Research: Atmospheres}, 113(D1), 2008.

\bibitem{Nebeker1995}
{Frederik Nebeker}.
\newblock {\em Calculating the weather: Meteorology in the 20th century.}
\newblock Elsevier, 1995.

\bibitem{friedman2001greedy}
Jerome~H Friedman.
\newblock Greedy function approximation: a gradient boosting machine.
\newblock {\em Annals of statistics}, pages 1189--1232, 2001.

\bibitem{funahashi1993approximation}
Ken-ichi Funahashi and Yuichi Nakamura.
\newblock Approximation of dynamical systems by continuous time recurrent
  neural networks.
\newblock {\em Neural networks}, 6(6):801--806, 1993.

\bibitem{gers1999learning}
Felix~A Gers, J{\"u}rgen Schmidhuber, and Fred Cummins.
\newblock Learning to forget: Continual prediction with lstm.
\newblock {\em Neural Computation}, 12(10):2451--2471, 2000.

\bibitem{gonccalves2017spatial}
Andre~R Goncalves, Arindam Banerjee, and Fernando~J Von~Zuben.
\newblock Spatial projection of multiple climate variables using hierarchical
  multitask learning.
\newblock In {\em Thirty-First AAAI Conference on Artificial Intelligence},
  2017.

\bibitem{gonccalves2014multi}
Andre~R Goncalves, Puja Das, Soumyadeep Chatterjee, Vidyashankar Sivakumar,
  Fernando~J Von~Zuben, and Arindam Banerjee.
\newblock Multi-task sparse structure learning.
\newblock In {\em Proceedings of the 23rd ACM International Conference on
  Conference on Information and Knowledge Management}, pages 451--460, 2014.

\bibitem{gonccalves2016multi}
Andre~R Goncalves, Fernando~J Von~Zuben, and Arindam Banerjee.
\newblock Multi-task sparse structure learning with gaussian copula models.
\newblock {\em The Journal of Machine Learning Research}, 17(1):1205--1234,
  2016.

\bibitem{grover2015deep}
Aditya Grover, Ashish Kapoor, and Eric Horvitz.
\newblock A deep hybrid model for weather forecasting.
\newblock In {\em Proceedings of the 21th ACM SIGKDD International Conference
  on Knowledge Discovery and Data Mining}, pages 379--386. ACM, 2015.

\bibitem{ham2019deep}
Yoo-Geun Ham, Jeong-Hwan Kim, and Jing-Jia Luo.
\newblock Deep learning for multi-year enso forecasts.
\newblock {\em Nature}, 573(7775):568--572, 2019.

\bibitem{he2019interpretable}
Sijie He, Xinyan Li, Vidyashankar Sivakumar, and Arindam Banerjee.
\newblock Interpretable predictive modeling for climate variables with weighted
  lasso.
\newblock In {\em Proceedings of the AAAI Conference on Artificial
  Intelligence}, volume~33, pages 1385--1392, 2019.

\bibitem{LSTM}
Sepp Hochreiter and J\"{u}rgen Schmidhuber.
\newblock Long short-term memory.
\newblock {\em Neural Comput.}, 9(8):1735--1780, November 1997.

\bibitem{soil_1}
Jin Huang, Huug~M van~den Dool, and Konstantine~P Georgarakos.
\newblock Analysis of model-calculated soil moisture over the united states
  (1931--1993) and applications to long-range temperature forecasts.
\newblock {\em Journal of Climate}, 9(6):1350--1362, 1996.

\bibitem{hwang2019improving}
Jessica Hwang, Paulo Orenstein, Judah Cohen, Karl Pfeiffer, and Lester Mackey.
\newblock Improving subseasonal forecasting in the western us with machine
  learning.
\newblock In {\em Proceedings of the 25th ACM SIGKDD International Conference
  on Knowledge Discovery \& Data Mining}, pages 2325--2335. ACM, 2019.

\bibitem{jalali2013dirty}
Ali Jalali, Pradeep Ravikumar, and Sujay Sanghavi.
\newblock A dirty model for multiple sparse regression.
\newblock {\em IEEE Transactions on Information Theory}, 59(12):7947--7968,
  2013.

\bibitem{reanalysis}
E.~Kalnay, M.~Kanamitsu, R.~Kistler, W.~Collins, D.~Deaven, L.~Gandin,
  M.~Iredell, S.~Saha, G.~White, J.~Woollen, Y.~Zhu, M.~Chelliah, W.~Ebisuzaki,
  W.~Higgins, J.~Janowiak, K.~C. Mo, C.~Ropelewski, J.~Wang, A.~Leetmaa,
  R.~Reynolds, Roy Jenne, and Dennis Joseph.
\newblock The ncep/ncar 40-year reanalysis project.
\newblock {\em Bulletin of the American Meteorological Society},
  77(3):437--472, 1996.

\bibitem{klemm2017development}
Toni Klemm and Renee~A McPherson.
\newblock The development of seasonal climate forecasting for agricultural
  producers.
\newblock {\em Agricultural and forest meteorology}, 232:384--399, 2017.

\bibitem{kunsch1989jackknife}
Hans~R Kunsch.
\newblock The jackknife and the bootstrap for general stationary observations.
\newblock {\em The annals of Statistics}, pages 1217--1241, 1989.

\bibitem{lecun1998gradient}
Yann LeCun, L{\'e}on Bottou, Yoshua Bengio, and Patrick Haffner.
\newblock Gradient-based learning applied to document recognition.
\newblock {\em Proceedings of the IEEE}, 86(11):2278--2324, 1998.

\bibitem{lipovetsky2001shapley}
Stan Lipovetsky and Michael Conklin.
\newblock Analysis of regression in game theory approach.
\newblock {\em Applied Stochastic Models in Business and Industry},
  17(4):319--330, 2001.

\bibitem{lorenc1986analysis}
Andrew~C Lorenc.
\newblock Analysis methods for numerical weather prediction.
\newblock {\em Quarterly Journal of the Royal Meteorological Society},
  112(474):1177--1194, 1986.

\bibitem{lorenz1956empirical}
Edward~N Lorenz.
\newblock Empirical orthogonal functions and statistical weather prediction.
\newblock 1956.

\bibitem{boar16}
{National Academies of Sciences}.
\newblock {\em Next generation earth system prediction: strategies for
  subseasonal to seasonal forecasts}.
\newblock National Academies Press, 2016.

\bibitem{nati10}
{National Research Council}.
\newblock {\em Assessment of intraseasonal to interannual climate prediction
  and predictability}.
\newblock National Academies Press, 2010.

\bibitem{pomeroy2002prediction}
JW~Pomeroy, DM~Gray, NR~Hedstrom, and JR~Janowicz.
\newblock Prediction of seasonal snow accumulation in cold climate forecasts.
\newblock {\em Hydrological Processes}, 16(18):3543--3558, 2002.

\bibitem{radhika2009atmospheric}
Y~Radhika and M~Shashi.
\newblock Atmospheric temperature prediction using support vector machines.
\newblock {\em International journal of computer theory and engineering},
  1(1):55, 2009.

\bibitem{reichstein2019deep}
Markus Reichstein, Gustau Camps-Valls, Bjorn Stevens, Martin Jung, Joachim
  Denzler, Nuno Carvalhais, et~al.
\newblock Deep learning and process understanding for data-driven earth system
  science.
\newblock {\em Nature}, 566(7743):195--204, 2019.

\bibitem{reshef2011detecting}
David~N Reshef, Yakir~A Reshef, Hilary~K Finucane, Sharon~R Grossman, Gilean
  McVean, Peter~J Turnbaugh, Eric~S Lander, Michael Mitzenmacher, and Pardis~C
  Sabeti.
\newblock Detecting novel associations in large data sets.
\newblock {\em science}, 334(6062):1518--1524, 2011.

\bibitem{reynolds2007daily}
Richard~W Reynolds, Thomas~M Smith, Chunying Liu, Dudley~B Chelton, Kenneth~S
  Casey, and Michael~G Schlax.
\newblock Daily high-resolution-blended analyses for sea surface temperature.
\newblock {\em Journal of Climate}, 20(22):5473--5496, 2007.

\bibitem{schneider2017earth}
Tapio Schneider, Shiwei Lan, Andrew Stuart, and Joao Teixeira.
\newblock Earth system modeling 2.0: A blueprint for models that learn from
  observations and targeted high-resolution simulations.
\newblock {\em Geophysical Research Letters}, 44(24):12--396, 2017.

\bibitem{shi2017deep}
Xingjian Shi, Zhihan Gao, Leonard Lausen, Hao Wang, Dit-Yan Yeung, Wai-kin
  Wong, and Wang-chun Woo.
\newblock Deep learning for precipitation nowcasting: A benchmark and a new
  model.
\newblock In {\em Advances in neural information processing systems}, pages
  5617--5627, 2017.

\bibitem{simmholl2002}
A.~J. Simmons and A.~Hollingsworth.
\newblock Some aspects of the improvement in skill of numerical weather
  prediction.
\newblock {\em Quarterly Journal of the Royal Meteorological Society},
  128(580):647--677, 2002.

\bibitem{simmons2002some}
Adrian~J Simmons and Anthony Hollingsworth.
\newblock Some aspects of the improvement in skill of numerical weather
  prediction.
\newblock {\em Quarterly Journal of the Royal Meteorological Society: A journal
  of the atmospheric sciences, applied meteorology and physical oceanography},
  128(580):647--677, 2002.

\bibitem{srivastava2015unsupervised}
Nitish Srivastava, Elman Mansimov, and Ruslan Salakhudinov.
\newblock Unsupervised learning of video representations using lstms.
\newblock In {\em International conference on machine learning}, pages
  843--852, 2015.

\bibitem{sundermeyer2012lstm}
Martin Sundermeyer, Ralf Schl{\"u}ter, and Hermann Ney.
\newblock Lstm neural networks for language modeling.
\newblock In {\em Thirteenth annual conference of the international speech
  communication association}, 2012.

\bibitem{sutskever2014sequence}
Ilya Sutskever, Oriol Vinyals, and Quoc~V Le.
\newblock Sequence to sequence learning with neural networks.
\newblock In {\em Advances in neural information processing systems}, pages
  3104--3112, 2014.

\bibitem{predictability_s2s}
{The National Oceanic and Atmospheric Administration}.
\newblock {\em Subseasonal and Seasonal Forecasting Innovation: Plans for the
  Twenty-First Century}.
\newblock Annotated outline of NOAA's draft sub-seasonal and seasonal (S2S)
  forecasting report, 2018.

\bibitem{lasso}
Robert Tibshirani.
\newblock Regression shrinkage and selection via the lasso.
\newblock {\em Journal of the Royal Statistical Society,}, pages 267--288,
  1996.

\bibitem{NAO_2}
HM~Van~den Dool, S~Saha, and AAke Johansson.
\newblock Empirical orthogonal teleconnections.
\newblock {\em Journal of Climate}, 13(8):1421--1435, 2000.

\bibitem{van2007empirical}
Huug Van~den Dool, Principal~Scientist Cpc, and Huug Van Den~Dool.
\newblock {\em Empirical methods in short-term climate prediction}.
\newblock Oxford University Press, 2007.

\bibitem{soil_2}
Huug Van~den Dool, Jin Huang, and Yun Fan.
\newblock Performance and analysis of the constructed analogue method applied
  to us soil moisture over 1981--2001.
\newblock {\em Journal of Geophysical Research: Atmospheres}, 108(D16), 2003.

\bibitem{vaswani2017attention}
Ashish Vaswani, Noam Shazeer, Niki Parmar, Jakob Uszkoreit, Llion Jones,
  Aidan~N Gomez, {\L}ukasz Kaiser, and Illia Polosukhin.
\newblock Attention is all you need.
\newblock In {\em Advances in neural information processing systems}, pages
  5998--6008, 2017.

\bibitem{venugopalan2015sequence}
Subhashini Venugopalan, Marcus Rohrbach, Jeffrey Donahue, Raymond Mooney,
  Trevor Darrell, and Kate Saenko.
\newblock Sequence to sequence-video to text.
\newblock In {\em Proceedings of the IEEE international conference on computer
  vision}, pages 4534--4542, 2015.

\bibitem{vitart2012}
Frédéric Vitart, Andrew~W Robertson, and David~LT Anderson.
\newblock Subseasonal to seasonal prediction project: Bridging the gap between
  weather and climate.
\newblock {\em Bulletin of the World Meteorological Organization}, 61(23),
  2012.

\bibitem{von2001statistical}
Hans Von~Storch and Francis~W Zwiers.
\newblock {\em Statistical analysis in climate research}.
\newblock Cambridge university press, 2001.

\bibitem{wasserman2013all}
Larry Wasserman.
\newblock {\em All of statistics: a concise course in statistical inference}.
\newblock Springer Science \& Business Media, 2013.

\bibitem{weisberg2005applied}
Sanford Weisberg.
\newblock {\em Applied linear regression}, volume 528.
\newblock John Wiley \& Sons, 2005.

\bibitem{MJO}
Matthew~C. Wheeler and Harry~H. Hendon.
\newblock An all-season real-time multivariate mjo index: Development of an
  index for monitoring and prediction.
\newblock {\em Monthly Weather Review}, 132(8):1917--1932, 2004.

\bibitem{xingjian2015convolutional}
Shi Xingjian, Zhourong Chen, Hao Wang, Dit-Yan Yeung, Wai-Kin Wong, and
  Wang-chun Woo.
\newblock Convolutional lstm network: A machine learning approach for
  precipitation nowcasting.
\newblock In {\em Advances in neural information processing systems}, pages
  802--810, 2015.

\bibitem{zimmerman2016utilizing}
Brian~G Zimmerman, Daniel~J Vimont, and Paul~J Block.
\newblock Utilizing the state of enso as a means for season-ahead predictor
  selection.
\newblock {\em Water resources research}, 52(5):3761--3774, 2016.

\end{thebibliography}

\newpage
\clearpage
\appendix
\section{Difficulty of the Problem}

\label{sec:app_ssf_difficulty}

\subsection{Dependence between historical data and forecasting target}
\begin{figure}[h]
\centering
\includegraphics[width=\textwidth]{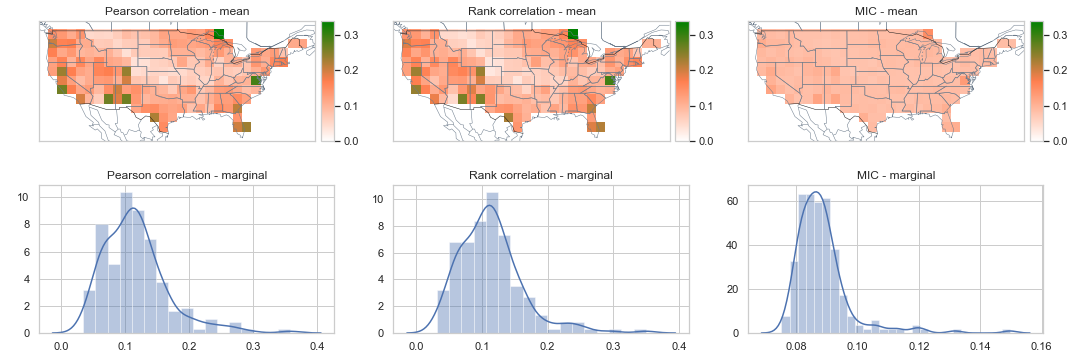}
\caption{Pearson correlation, Spearman's rank correlation and MIC between 2m temperature of week -2 \& -1 and week 3 \& 4. Small values ($\le$0.2) of Pearson correlation and Spearman's rank correlation at a majority of locations verify the fact, as we illustrate in the main paper using MIC, that there is little information shared between the most recent date and the forecasting target.}
\label{fig:corr_margin}
\end{figure}

In section \ref{sec:ssf_difficulty}, the dependence between the most recent historical data (the normalized average temperature of week -2 \& -1) and the forecasting target (the normalized average temperature of week 3 \& 4) is measured by maximum information coefficient (MIC). Here we show the results measured by Pearson correlation coefficient \cite{wasserman2013all}, and Spearman's rank correlation coefficient \cite{wasserman2013all} (Figure \ref{fig:corr_margin}). Small values ($\le$0.2) of Pearson correlation and Spearman's rank correlation at a majority of locations, which verify that there is little information shared between the most recent date and the forecasting target, once again, demonstrate how difficult SSF is.

\subsection{Relative R$\mathbf{^2}$}

In the main paper, we introduce cosine similarity, which is widely used in weather prediction evaluation, as an evaluation metric. Here we formally define the other evaluation metric, namely relative $R^2$ as

\begin{equation}
\text{Relative } R^2=1-\text{Relative MSE}=
1-\frac{\sum_{i=1}^n(\y^*_i-\hat{\y}_i)^2}{\sum_{i=1}^n(\y^*_i-\bar{\y}_{\text{train}})^2}~,
\end{equation}

where $\hat{\y}$ denotes a vector of predicted values, and $\y^*$ be the corresponding ground truth. We use relative $R^2$ to evaluate the relative predictive skill of a given prediction $\hat{y}$ compared to the best constant predictor $\bar{y}_{\text{train}}$, the long-term average of target variable at each date and each target location computed from training set. A model which achieves a positive relative $R^2$ is, at least, able to predict the sign of $y^*$ accurately. The results of temporal and spatial relative $R^2$ over the US mainland of ML models discussed in section \ref{sec:models} are shown in Table \ref{tab:r2_all} and Figure \ref{fig:spatial_result_r2} respectively.

\section{Data and experimental setup}
\label{sec:app_exp_setup}
\subsection{Data sources}
The data described in Table \ref{table: data} were downloaded from the following sources:
\begin{itemize}[leftmargin=*]
    \item Temperature (tmp2m):
    \url{https://www.esrl.noaa.gov/psd/data/gridded/data.cpc.globaltemp.html}
    \item Soil moisture (sm): \url{https://www.esrl.noaa.gov/psd/data/gridded/data.cpcsoil.html}
    \item Sea surface temperature (sst): \url{https://www.ncdc.noaa.gov/oisst}
    \item Relative humidity (rhum), sea level pressure (slp), and geopotential height (hgt): \url{ftp://ftp.cdc.noaa.gov/Datasets/ncep.reanalysis/surface/}
    \item Multivariate ENSO index (MEI): \url{https://psl.noaa.gov/enso/mei/}
    \item Ni\~no indices: \url{https://www.cpc.ncep.noaa.gov/data/indices/wksst8110.for}
    \item North Atlantic Oscillation (NAO) index: \url{ftp://ftp.cpc.ncep.noaa.gov/cwlinks/norm.daily.nao.index.b500101.current.ascii}
    \item Madden Julian Oscillation (MJO) phase \& amplitude: \url{http://www.bom.gov.au/climate/mjo/graphics/rmm.74toRealtime.txt}
\end{itemize}

\subsection{PCA prepossessing}

As mentioned in section \ref{sec:exp_setup} of the main paper, one way for feature extraction is to apply PCA to spatial-temporal variables. To do so, let's consider sst of Pacific ocean as an example. Daily sst of Pacific ocean is originally stored in a matrix, of which each element represents the sea surface temperature at each grid point of Pacific ocean. The covariance matrix can be computed by flattening each matrix into a 1-D vector, viewing each element in the matrix as a feature and each date as one observation. Such covariance matrix captures spatial connection among grid points of Pacific ocean. By considering all dates from 1986 to 2016, we can extract the top 10 principal components (PCs) as features based on PC loadings computed from the corresponding covariance.

\begin{figure}[t]
    \centering
    \subfigure[Sequential feature set for Mar 1, 2018]{\includegraphics[width=0.45\columnwidth]{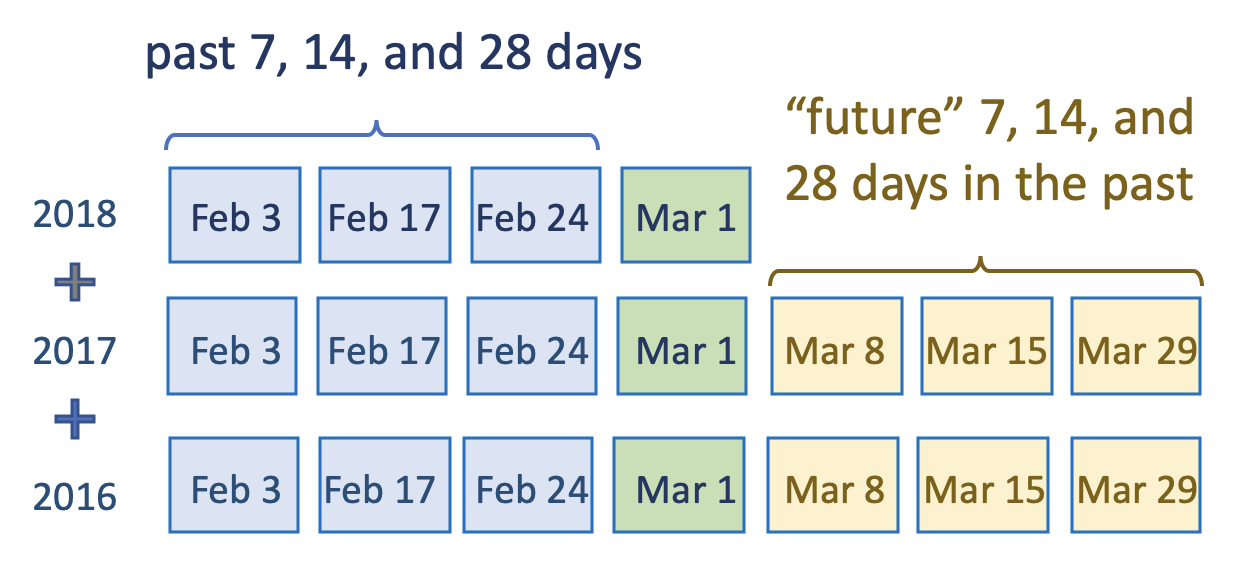}}
    \hspace{5mm}
    \subfigure[Evaluation pipeline]{\includegraphics[width=0.35\columnwidth]{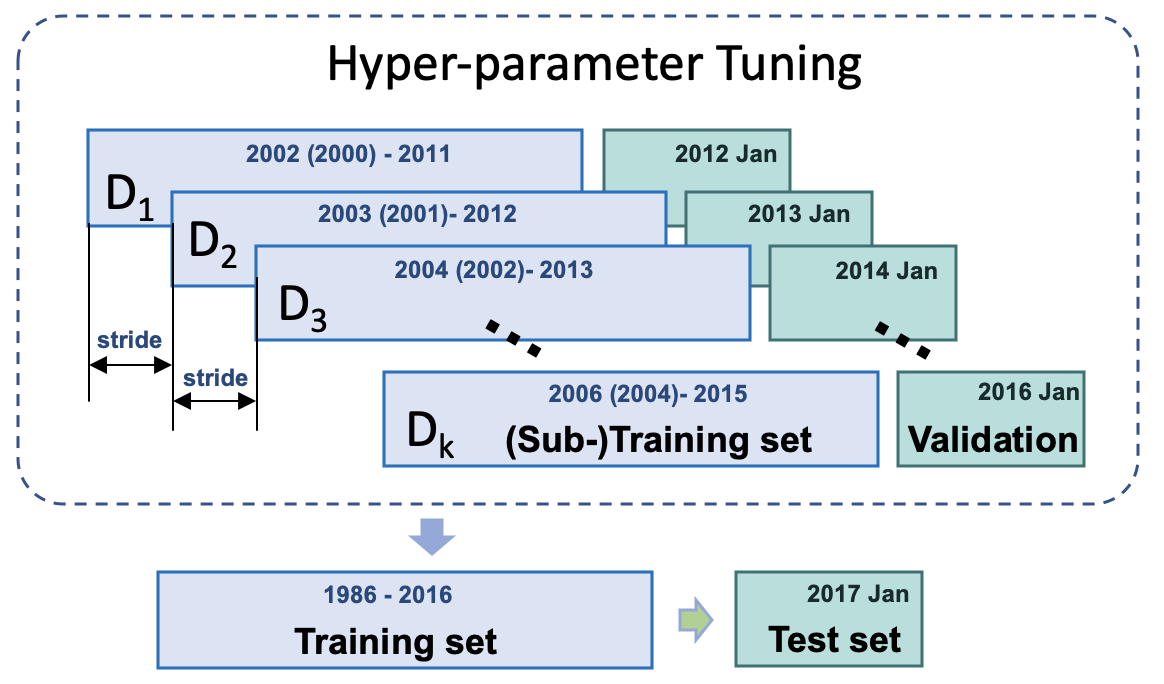}}
    \vspace{-2mm}
    \caption{(a) Sequential feature set: to construct feature set at Mar. 1, 2018, we concatenate covariates from Mar. 1 in 2018, 2017, and 2016, their corresponding $7^{th}$, $14^{th}$, and $28^{th}$ days in the past, and $7^{th}$, $14^{th}$, and $28^{th}$ ``future'' days in 2017 and 2016. (b) Evaluation pipeline: to test SSF in Jan 2017, 
    the training set covers historical 30 year ends at Dec 4, 2016 (the last available date). 5 validation sets include dates from each Jan between 2012 to 2016, with the corresponding training sets generated by applying a moving window of 10 years and a stride of 365 days on data start from 2000.}
     \label{fig:model_selection}
\end{figure}

 \subsection{Feature set construction}
 
To better utilize historical information, we construct a sequential feature set by including not only covariates of the target date, but also covariates of the $7^{th}$, $14^{th}$, and $28^{th}$ day previous from the target date, as well as the day of the year of the target date in the past 2 years and both the historical past and future dates around the day of the year of the target date in the past 2 years. Such selection of historical dates mainly bases on the temporal correlation. Figure \ref{fig:model_selection}(a) provides a detailed example on how to construct feature set for Mar 1, 2018: we concatenate covariates from Mar. 1 in 2018, 2017, and 2016, their corresponding $7^{th}$, $14^{th}$, and $28^{th}$ days in the past, and $7^{th}$, $14^{th}$, and $28^{th}$ ``future'' days in 2017 and 2016. In total, we include $H=18$ historical days in our feature set for each date.

\subsection{Evaluation pipeline}

Predictive models are created independently for each month in 2017 and 2018. To mimic a live forecasting system, we generate 105 test dates during 2017-2018, one for each week, and group them into 24 test sets by their month of the year. Given a test set, our evaluation pipeline consists of two parts (Figure~\ref{fig:model_selection}(b)):

\begin{itemize}[leftmargin=*]
\setlength\itemsep{0em}
\item  ``5-fold'' training-validation pairs for hyper-parameter tuning, based on a ``sliding-window'' strategy designed for time-series data. Each validation set uses the data from the same month of the year as the test set. For instance, if the test set is Jan 2017, the corresponding 5 validation sets are Jan 2012, Jan 2013, Jan 2014, Jan 2015, and Jan 2016 respectively. Each validation set corresponds to a training set containing 10 years of data and ending 28 days before the first date in the validation set. Specifically, if the validation set starting from Jan 1, 2016, 
the training set is from Dec 4, 2005 to Dec 4, 2015. Such construction is equivalent to apply a sliding-window of 10-year with a stride of 365 days on data from 2002.

\item The training-test pair, where the training set, including 30-year data in the past, ends 28 days before the first date in the test set. For example, to test SSF in Jan 2017, i.e., Jan 1, Jan 8, Jan 15, Jan 22, and Jan 29, the training set starts from Dec 4, 1986 and ends at Dec 4, 2016, which is the $28^{th}$ day before Jan 1, and the last date we have the ground truth for the target variable.
\end{itemize}

\section{Additional Results}
\label{sec:app_results}

\begin{table}
  \caption{Comparison of relative $R^2$ of tmp2m forecasting for test sets over 2017-2018. A positive relative $R^2$ indicates a model predicting the sign of the target variable correctly. XGBoost achieves the highest relative $R^2$. 
  }

  \resizebox{\textwidth}{!}{
  \begin{tabular}{c|c c c c}
  \hline
    Model & Mean(se) & Median (se) & 0.25 quantile (se) & 0.75 quantile (se)\\
  \hline
  \multicolumn{5}{c}{\textbf{Temporally Global Set}}\\
\hline
    \textbf{XGBoost - one day }&\textbf{0.0760(0.03)} &\textbf{0.0974(0.03)}  &\textbf{-0.0449(0.03)}&\textbf{ 0.2434(0.03)}\\
  Lasso - one day &  0.0552(0.02) &0.0321(0.02)& -0.0309(0.01)& 0.1295(0.02)\\
    \hline
     \textbf{Encoder (LSTM)-Decoder (FNN)}&  \textbf{-0.0353 (0.05)} &\textbf{0.0596(0.05)}&\textbf{-0.2409 (0.06)} &\textbf{ 0.2426 (0.05)}\\
     FNN & -0.5777(0.29)& -0.0183(0.15)&  -0.0794(0.13)& 0.0213(0.13)\\
    CNN & -0.0564(0.03)& 0.0284(0.02) & -0.0266(0.02)& 0.0570(0.02)\\
     CNN-LSTM &  -0.1164(0.05) &0.0263(0.03) & -0.0862(0.03)& 0.0698(0.03)\\
    \hline
\textbf{LS with NAO \& all nino  - daily }& \textbf{0.0418(0.01)} 
&\textbf{0.0535(0.01)}  &\textbf{-0.0078(0.01)}  & \textbf{0.0949(0.01)} \\
Damped persistence & 0.0266(0.01) &0.0414(0.02) & -0.0542(0.02) & 0.1354(0.02)\\
    \hline 
    MultiLLR  &-0.0571 (0.02) & 0.0034 (0.02)
&-0.1156 (0.03) & 0.0797 (0.02)\\
AutoKNN  &0.0181 (0.01) & 0.0260 (0.02)&-0.0531 (0.02) & 0.1041 (0.01)\\
    \hline
\multicolumn{5}{c}{\textbf{Temporally Local Set}}\\
\hline
XGBoost - one day &-0.0337(0.03) &0.0396(0.03)&  -0.1310(0.04) &0.1873(0.03)\\
Lasso - one day &    -0.0028(0.02)& 0.0327(0.02) & -0.0613(0.02)& 0.0996(0.02)\\
Encoder (LSTM)-Decoder (FNN)&-0.2333 (0.06) & -0.1116 (0.06)
&-0.4694 (0.09) & 0.1808 (0.06)\\
\hline
\end{tabular}}
\label{tab:r2_all}
\end{table}

\begin{figure}[t]
\centering
\includegraphics[width=\textwidth]{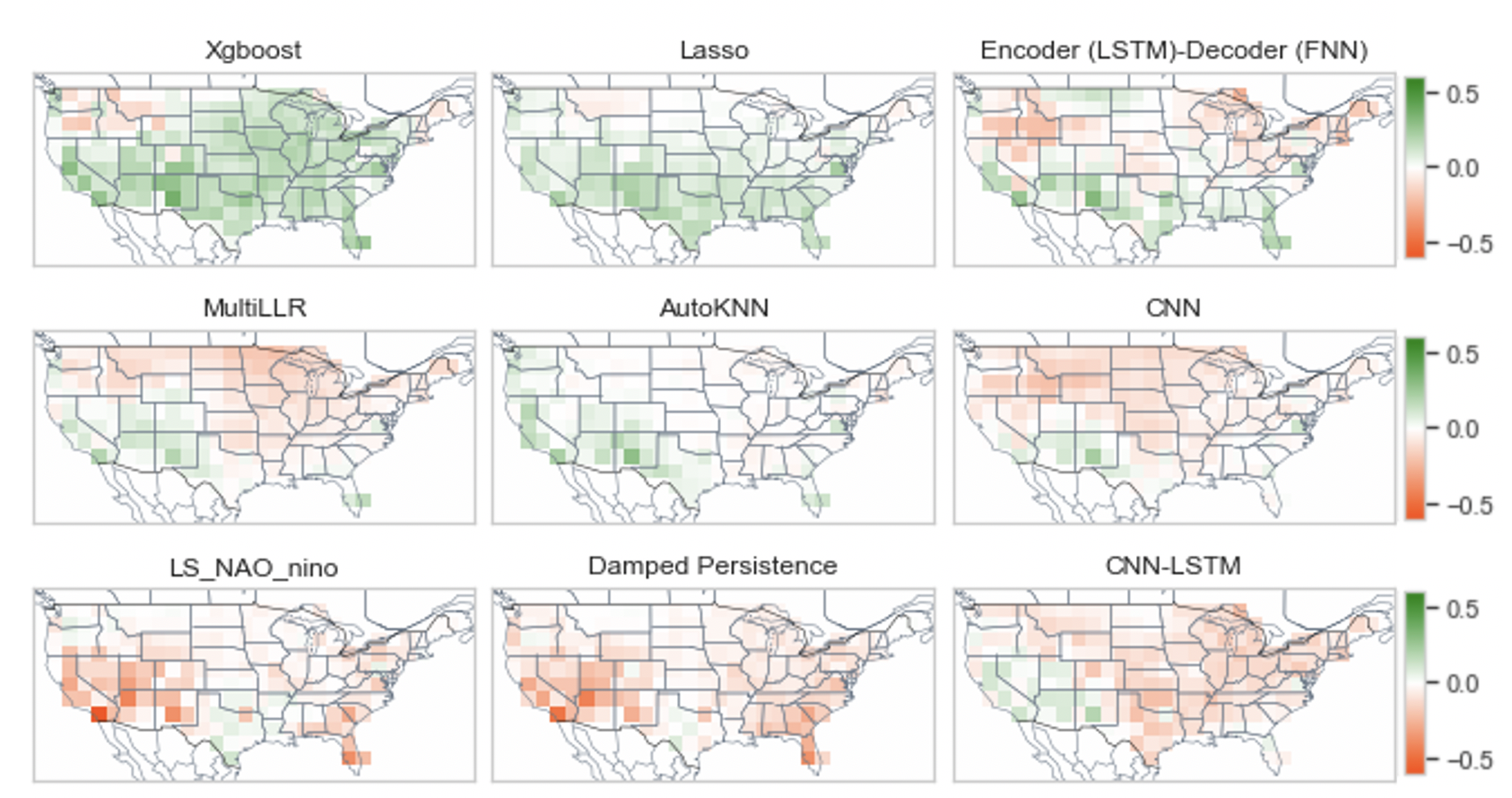}
\caption{Temporal relative $R^2$ over the US mainland of ML models discussed in section \ref{sec:models} for temperature prediction over 2017-2018. Large positive values (green) closer to 1 indicates better predictive skills.}
\label{fig:spatial_result_r2}
\end{figure}

\subsection{Temporal and spatial results of relative R$\mathbf{^2}$}

Table \ref{tab:r2_all} lists the mean, the median, the 0.25 quantile, the 0.75 quantile, and their corresponding standard errors of relative $R^2$ for all models.  A positive relative $R^2$ indicates a model can at least predict the sign of the target variable correctly. Again, XGBoost achieves the highest predictive skill in terms of both the mean and the median, demonstrating its predictive power. Linear regression, like Lasso, with a proper feature set has good predictive performance. Both XGBoost and Lasso have larger positive relative $R^2$ in terms of the mean, and can still outperform climatology and two climate baseline models, i.e., LS with NAO \& Ni\~no, and damped persistence. Even though Encoder (LSTM)-Decoder (FNN) has a slightly negative mean relative $R^2$, it has the second largest median and 0.75 quantile among all models, showing its potential for further improvement. 

Figure \ref{fig:spatial_result_r2} shows the spatial relative $R^2$ of all methods. XGBoost and Lasso are able to achieve positive relative $R^2$ for most of the target locations. Encoder (LSTM)-Decoder (FNN) shows better predictive skill over the southern US compared to other regions. MultiLLR and AutoKNN manages to obtain non-negative relative $R^2$ for the coastal area in the western US but their predictive performance drops in the rest of locations. All other baseline methods struggle to reach positive relative $R^2$ for most of the target locations.

\subsection{Analysis on feature importance}

To emphasis the importance of the land-based covariates, e.g., soil moisture and the ocean-based covariates, e.g., NAO and Ni\~no indices, we compare the prediction performance among (1) the model trained with all covariates, (2) the model trained without soil moisture, and (3) the model trained without NAO and Ni\~no indices (Table~\ref{tab:cos_without} and Table~\ref{tab:r2_without}). Most models experience a performance deterioration when we exclude certain ``important'' covariates.

\subsection{The influence of feature sequence length} We compare the prediction performance under 3 different settings, referred to as ``one day'', ``four days", and ``all days'' respectively. For feature set construction, ``one day'' includes covariates at the target date only, ``four days'' also covers the  $7^{th}$, $14^{th}$, and $28^{th}$ days previous to the target date, and ``all days'' uses the exact feature sequence we use for LSTM-based models. Comparison of predictive skills under each setting, measured by both cosine similarity and relative $R^2$, can be found in Table~\ref{tab:cos_seq_len} and Table~\ref{tab:r2_seq_len}. Both XGBoost and Lasso enjoy a performance boost using ``one day'' values. Especially for XGBoost, the performance of ``one day'' is approximately 50\% better than using ``all days''. A possible explanation for such performance degradation as we increase the feature sequence length is that both models weight covariates from different dates exactly the same without considering temporal information, thus more noise has been introduced. 

\begin{table}[t]
  \caption{Comparison of cosine similarity of tmp2m forecasting for test sets over 2017-2018 using different feature set. Excluding soil moisture or climate indices (NAO \& Ni\~no) leads to a deterioration in the predictive performance.
  }

  \resizebox{\textwidth}{!}{
  \begin{tabular}{c|c c c c}
  \hline
    Model & Mean(se) & Median (se) & 0.25 quantile (se) & 0.75 quantile (se)\\
\hline
\textbf{XGBoost - one day}&\textbf{ 0.3044(0.03)}& \textbf{0.3447(0.05)}&  \textbf{0.0252(0.05)}&\textbf{ 0.5905(0.04})\\
XGBoost - one day (w/o soil moisture)&0.2685(0.03)& 0.2797(0.05) & 0.0703(0.04)& 0.5492(0.05)\\
XGBoost - one day (w/o nao \& all nino) & 0.2081(0.03)& 0.1640(0.05) & -0.0588(0.04) &0.5246(0.05)\\
\hline
 Lasso - one day &  0.2499(0.04)& 0.2554(0.06) & -0.0224(0.05) &0.5604(0.06)\\
 Lasso - one day (w/o soil moisture)&0.2638(0.04)& 0.2912(0.05) & 0.0032(0.06)& 0.5655(0.05)\\
 Lasso - one day (w/o nao \& all nino)& 0.1956(0.04)& 0.2573(0.07) & -0.1657(0.06) &0.5533(0.05)\\
\hline
Encoder (LSTM)-Decoder (FNN) & 0.2616 (0.04) & 0.2995 (0.07)
&{-0.0719 (0.06)} & {0.6310 (0.05)}\\
Encoder (LSTM)-Decoder (FNN)(w/o soil moisture)&0.2157 (0.04) & 0.2909 (0.07)
&-0.1106 (0.07) & 0.5443 (0.07)\\

Encoder (LSTM)-Decoder (FNN)(w/o nao \& all nino)
&0.2236 (0.04) & 0.2395 (0.06)
&-0.1527 (0.07) & 0.5989 (0.06)\\
\hline 
\end{tabular}}
\label{tab:cos_without}
\end{table}

\begin{table}[th]
  \caption{Comparison of relative $R^2$ of tmp2m forecasting for test sets over 2017-2018. Excluding soil moisture or climate indices (NAO \& Ni\~no) leads to a smaller or even negative relative $R^2$, showing that it becomes harder for the model to predict the sign of the target variable correctly.}
 
  \resizebox{\textwidth}{!}{
  \begin{tabular}{c|c c c c}
  \hline
    Model & Mean(se) & Median (se) & 0.25 quantile (se) & 0.75 quantile (se)\\
\hline
\textbf{XGBoost - one day }&\textbf{0.0760(0.03)} &\textbf{0.0974(0.03)}  &\textbf{-0.0449(0.03)}&\textbf{ 0.2434(0.03)}\\
XGBoost - one day (w/o soil moisture)&0.0370(0.03)& 0.0322(0.03) & -0.0564(0.03) &0.2225(0.03)\\
XGBoost - one day (w/o nao \& all nino)&-0.0161(0.03)& -0.0079(0.04) & -0.1618(0.03)& 0.2426(0.04)\\
\hline
Lasso - one day &  0.0552(0.02) &0.0321(0.02)& -0.0309(0.01)& 0.1295(0.02)\\
Lasso - one day (w/o soil moisture)&-0.0161(0.03)& -0.0079(0.04) & -0.1618(0.03) &0.2426(0.04)\\
Lasso - one day (w/o nao \& all nino)&0.0003(0.02)& 0.0457(0.02) & -0.1113(0.03)& 0.1641(0.02)\\
\hline
Encoder (LSTM)-Decoder (FNN) &  -0.0353 (0.05) &0.0596(0.05)&-0.2409 (0.06) &0.2426 (0.05)\\
Encoder (LSTM)-Decoder (FNN)(w/o soil moisture)
&-0.1083 (0.05) & 0.0314 (0.05)
&-0.3022 (0.08) & 0.2252 (0.05)\\

Encoder (LSTM)-Decoder (FNN)(w/o nao \& all nino)
&-0.0802 (0.04) & 0.0124 (0.05)
&-0.3032 (0.06) & 0.2446 (0.05)\\
\hline 
\end{tabular}}
\label{tab:r2_without}
\end{table}

\begin{table}[t]
   \caption{Comparison of spatial cosine similarity for tmp2m forecasting over 2017-2018 using various length of feature sequence. Including longer historical sequence leads to a deterioration in the predictive performance of XGBoost and Lasso.}
 
   \centering
   \resizebox{0.8\textwidth}{!}{
   \begin{tabular}{c|c c c c}
   \hline
   Model & Mean(se) & Median (se) & 0.25 quantile (se) & 0.75 quantile (se)\\
     \hline
     XGBoost - all days  &0.2080(0.03) &0.1582(0.05) & -0.0466(0.05)& 0.5383(0.05)\\
     XGBoost - four days  &0.2433(0.03)&0.2203(0.05) & 0.0561(0.04)& 0.5168(0.06)\\
    \textbf{XGBoost - one day }&\textbf{ 0.3044(0.03)}& \textbf{0.3447(0.05)}&  \textbf{0.0252(0.05)}&\textbf{ 0.5905(0.04})\\
     \hline
     Lasso - all days  &0.2160(0.04)&0.2258(0.07)&-0.1381(0.06)&0.5384(0.06)\\
     Lasso - four days &0.2247(0.04)& 0.1952(0.07)&0.0572(0.06)&-0.5700(0.06)\\
    \textbf{ Lasso - one day }&  \textbf{0.2499(0.04)}& \textbf{0.2554(0.06)} & \textbf{-0.0224(0.05}) &\textbf{0.5604(0.06)}\\
     \hline
     \end{tabular}
     }
     \label{tab:cos_seq_len}
\end{table}

\begin{table}[t]
  \caption{Comparison of relative $R^2$ (with training set mean) for tmp2m prediction for test set over 2017-2019 using different length of feature sequence. Including longer historical sequence leads to a smaller or even negative relative $R^2$ for both XGBoost and Lasso.}

   \centering
   \resizebox{0.8\textwidth}{!}{
   \begin{tabular}{c|c c c c}
   \hline
  Model & Mean(se) & Median (se) & 0.25 quantile (se) & 0.75 quantile (se)\\
     \hline
     XGBoost - all days  &-0.0200(0.03)& -0.0010(0.04)&  -0.1499(0.04)& 0.2304(0.04)\\
     XGBoost - four days  &0.0242(0.03)& 0.0193(0.03)&  -0.0786(0.03)& 0.1882(0.04)\\
     \textbf{XGBoost - one day }&\textbf{0.0760(0.03)} &\textbf{0.0974(0.03)}  &\textbf{-0.0449(0.03)}&\textbf{ 0.2434(0.03)}\\
     \hline
     Lasso - all days  &-0.0167(0.03)& 0.0367(0.03)&  -0.0639(0.02)& 0.1588(0.03)\\
     Lasso - four days &0.0518(0.02) &0.0266(0.02) & -0.0542(0.02)& 0.1653(0.03)\\
     \textbf{Lasso - one day} &  \textbf{0.0552(0.02}) &\textbf{0.0321(0.02)}&\textbf{ -0.0309(0.01)}& \textbf{0.1295(0.02)}\\
     \hline
     \end{tabular}
     }
     \label{tab:r2_seq_len}
\end{table}

\subsection{Discussion on deep learning models}

{\bf Results of DL models.} Table \ref{tab:cos_deep} and Table \ref{tab:r2_deep} compare the predictive skills of 5 DL models discussed in section~\ref{sec:results}, measured by both cosine similarity and relative $R^2$. Significant improvements can been observed as we evolve from the standard Encoder (LSTM)-Decoder (LSTM), to Encoder (LSTM)-Decoder (FNN)-last step, where ``last step'' indicates that FNN Decoder only uses the last step of the output sequence from LSTM Encoder, and finally to Encoder (LSTM)-Decoder (FNN) with FNN Decoder uses every step of the output sequence from LSTM Encoder.

One issue with Encoder (LSTM)-Decoder (FNN) is that the input features are shared by all target locations, which requires the model to identify the useful information for each locations without any help from the input.

{\bf Autoregressive (AR) component.}  Currently, the Encoder(LSTM)-Decoder(FNN) clearly considers climate covariates on a global scale, which are shared by all target locations. Nevertheless, SSF depends on not only global climate condition but also local weather change. Therefore, we seek a way to improve the model by adding an autoregressive (AR) component to capture the ``local'' information from historical data. We consider two variants of Encoder (LSTM)-Decoder (FNN). The first variant contains an AR component with the input as historical temperature at each target location, denoted as Encoder (LSTM)-Decoder (FNN)+AR. The second one includes both historical temperature and historical temporal climate variables, i.e., climate indices, as input features, denoted as Encoder (LSTM)-Decoder (FNN)+AR (CI). For both models, the final forecast is computed as a linear combination of the prediction from Encoder (LSTM)-Decoder (FNN) and the prediction from AR component for each location. Unexpectedly, as shown in Table \ref{tab:cos_deep} and Table \ref{tab:r2_deep}, simply adding the AR component to our Encoder(LSTM)-Decoder(FNN) does not help the model to perform better. However, we believe there is a better way to involve local information, and such modification is a promising direction that worth investigation in the future.

\begin{table}[t]
  \caption{Comparison of cosine similarity of tmp2m forecasting for test sets over 2017-2018 using different deep learning architectures. 
  }
  
  \resizebox{\textwidth}{!}{
  \begin{tabular}{c|c c c c}
  \hline
    Model & Mean(se) & Median (se) & 0.25 quantile (se) & 0.75 quantile (se)\\
\hline
Encoder (LSTM)-Decoder (LSTM) & 0.0740(0.03) &0.0358(0.04) & -0.1569(0.03) &0.2584(0.04)\\
Encoder (LSTM)-Decoder (FNN)-last step &0.1614 (0.05)
& 0.2061 (0.08) &-0.2590 (0.08) & 0.5720 (0.08)\\
\textbf{Encoder (LSTM)-Decoder (FNN)} &\textbf{ 0.2616 (0.04}) & \textbf{0.2995 (0.07)}
&\textbf{-0.0719 (0.06)} & \textbf{0.6310 (0.05)}\\
Encoder (LSTM)-Decoder (FNN)+AR &0.1733 (0.04) & 0.1922 (0.06) &-0.0863 (0.07) & 0.5225 (0.06)\\
Encoder (LSTM)-Decoder (FNN)+AR (CI) & 0.1852 (0.04) & 0.1986 (0.05) &-0.0838 (0.06) & 0.5164 (0.05)\\
\hline
\end{tabular}}
\label{tab:cos_deep}
\end{table}

\begin{table}[t]
  \caption{Comparison of relative $R^2$ of tmp2m forecasting for test sets over 2017-2018. A positive relative $R^2$ indicates a model predicting the sign of the target variable correctly.  
  }

  \resizebox{\textwidth}{!}{
  \begin{tabular}{c|c c c c}
  \hline
    Model & Mean(se) & Median (se) & 0.25 quantile (se) & 0.75 quantile (se)\\
\hline
Encoder (LSTM)-Decoder (LSTM) & -0.3947(0.05)& -0.2999(0.05) & -0.6606(0.08) &-0.0537(0.05)\\
Encoder (LSTM)-Decoder (FNN)-last step &-0.1709 (0.06) & 0.0217 (0.06)
&-0.4569 (0.11) & 0.2278 (0.06)\\
\textbf{Encoder (LSTM)-Decoder (FNN)}&  \textbf{-0.0353 (0.05)} &\textbf{0.0596(0.05)}&\textbf{-0.2409 (0.06)} &\textbf{ 0.2426 (0.05)}\\
Encoder (LSTM)-Decoder (FNN)+AR &-0.0414 (0.04) & -0.0041 (0.05) &-0.3027 (0.07) & 0.2309 (0.05))\\
Encoder (LSTM)-Decoder (FNN)+AR (CI) &-0.0563 (0.03) & -0.0380 (0.05) &-0.2365 (0.05) & 0.1951 (0.04)\\
\hline
\end{tabular}}
\label{tab:r2_deep}
\end{table}

\end{document}